\documentclass{article} 
\usepackage{iclr2025_conference,times}

\usepackage{amsmath,amsfonts,bm}

\def\eqref#1{equation~\ref{#1}}

\def\1{\bm{1}}

\DeclareMathAlphabet{\mathsfit}{\encodingdefault}{\sfdefault}{m}{sl}
\SetMathAlphabet{\mathsfit}{bold}{\encodingdefault}{\sfdefault}{bx}{n}

\usepackage{hyperref}
\usepackage{url}
\usepackage{subfigure}
\usepackage[T1]{fontenc}
\usepackage{xcolor}
\usepackage{lipsum}
\usepackage{graphicx}
\usepackage{multicol,multirow}
\usepackage{amsmath,amssymb,amsfonts}
\usepackage{mathrsfs}
\usepackage{amsthm}
\usepackage{booktabs}

\title{Dhoroni: Exploring Bengali Climate Change and Environmental Views with a Multi-Perspective News Dataset and Natural Language Processing}

\author{Azmine Toushik Wasi\thanks{Corresponding Author}, Wahid Faisal, Taj Ahmad, Abdur Rahman \thanks{Azmine Toushik Wasi and Wahid Faisal contributed to the work equally.}  \\
Department of Industrial and Production Engineering\\
Shahjalal University of Science and Technology\\
Sylhet, Bangladesh 3114\\
\texttt{\{azmine32,wahid56,tajahmad29,abdur37\}@student.sust.edu} \\
\And
Mst Rafia Islam \\
Department of Law \\
Independent University \\
Dhaka, Bangladesh 1212 \\
\texttt{2030391@iub.edu.bd}
}

\iclrfinalcopy 
\begin{document}

\maketitle

\begin{abstract}
Climate change poses critical challenges globally, disproportionately affecting low-income countries that often lack resources and linguistic representation on the international stage. Despite Bangladesh's status as one of the most vulnerable nations to climate impacts, research gaps persist in Bengali-language studies related to climate change and NLP. To address this disparity, we introduce Dhoroni, a novel Bengali (Bangla) climate change and environmental news dataset, comprising a 2300 annotated Bangla news articles, offering multiple perspectives such as political influence, scientific / statistical data, authenticity, position detection, and stakeholder involvement. Furthermore, we present an in-depth exploratory analysis of Dhoroni and introduce BanglaBERT-Dhoroni family, a novel baseline model family for climate and environmental opinion detection in Bangla, fine-tuned on our dataset. This research contributes significantly to enhancing accessibility and analysis of climate discourse in Bengali (Bangla), addressing crucial communication and research gaps in climate-impacted regions like Bangladesh with 180 million people.
\end{abstract}

\section{Introduction} \label{sec:intro}
Climate change, defined as the long-term alteration in Earth's climate patterns largely due to human activities, has inflicted extensive economic costs globally. The burning of fossil fuels, deforestation, and industrial activities have led to increased greenhouse gas emissions, accelerating global warming and disrupting natural ecosystems. These consequences are particularly severe in regions already grappling with socio-economic challenges. Low-income countries, especially in tropical regions, have borne the brunt of these disruptions, suffering disproportionately from rising sea levels, extreme weather events, droughts, and other environmental shifts. Unlike the main polluting nations, these countries contribute only minimally to global carbon emissions, yet face the most devastating consequences \citep{Naddaf2022,Abbass2022}. 

One of the key reasons behind this disproportionate impact is the limited capacity of these nations to adapt to rapid environmental changes. Many low-income countries lack the infrastructure and financial resources required to mitigate the effects of climate change. This results in exacerbated economic losses, with sectors like agriculture, fisheries, and tourism—critical sources of livelihood for these communities—being severely affected. Rising sea levels have led to the loss of arable land and displacement of coastal populations, while extreme weather events, such as cyclones and floods, have destroyed homes, infrastructure, and essential services \citep{Abbass2022}. 
These repeated disruptions make it harder for these countries to recover and rebuild, creating a cycle of vulnerability and increasing poverty. Moreover, these economic and environmental challenges are compounded by political instability and governance issues, making it difficult to implement effective climate policies or develop adaptive strategies. In contrast, wealthier, industrialized nations, which are the primary contributors to greenhouse gas emissions, are better equipped to cope with climate change through technological advancements, disaster management systems, and financial resources. As a result, the gap between those most responsible for climate change and those most affected continues to widen \citep{Naddaf2022}.
These impacts are not only persisting but escalating rapidly, with each passing year bringing more intense storms, longer droughts, and erratic climate patterns that affect vulnerable communities. As sea levels rise and weather patterns become more volatile, these countries will continue to face mounting challenges in maintaining food security, health, and overall economic stability \citep{Tollefson2022}. Without urgent global cooperation and targeted interventions, the existing inequalities in climate vulnerability and economic resilience are likely to deepen, pushing millions more into poverty and jeopardizing sustainable development efforts \cite{Olausson2009}.

\begin{figure}[t] 
\centering {\includegraphics[scale=0.4]{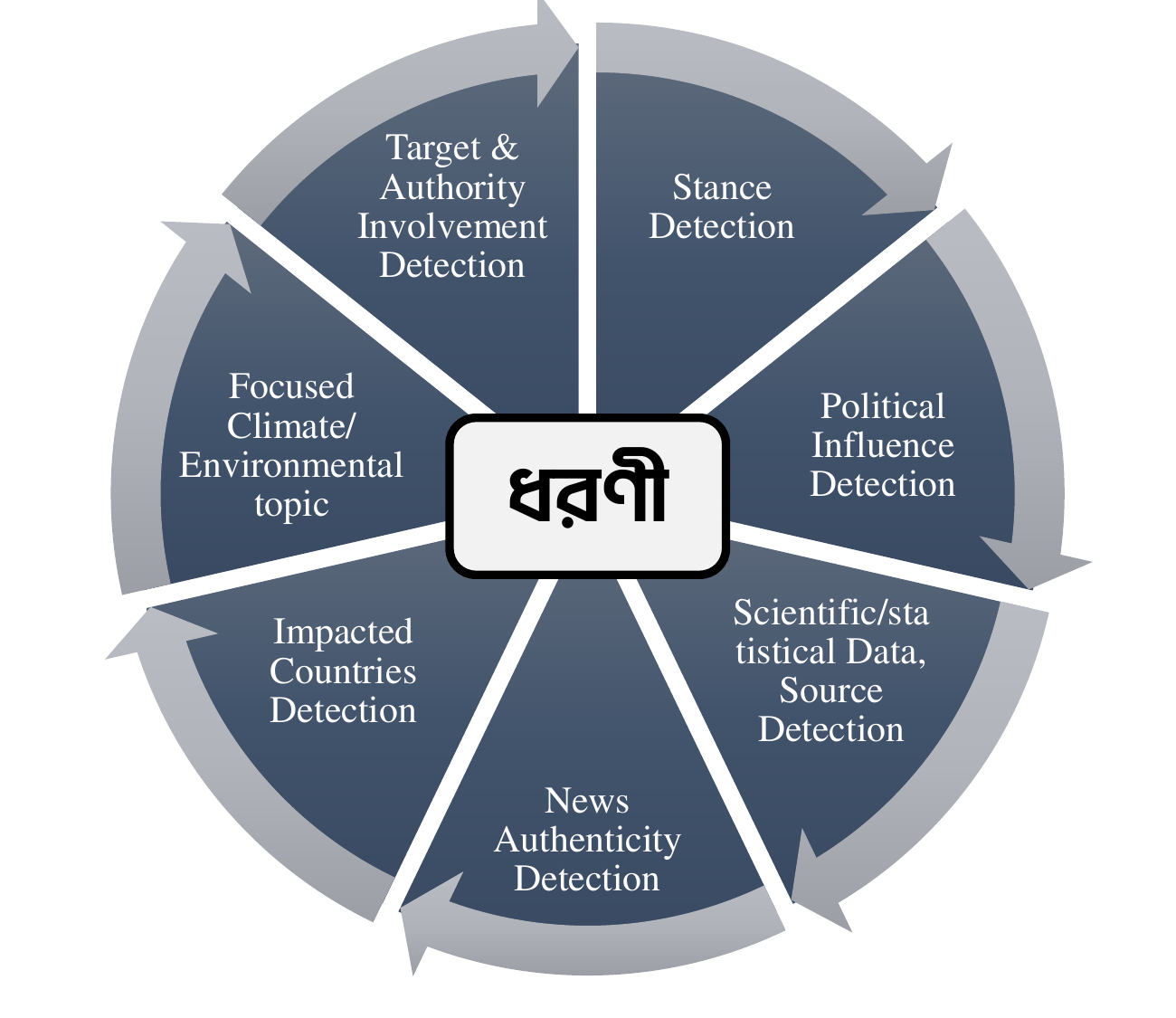}}
\caption{Multiple perspectives annotated and discussed in \texttt{Dhoroni} dataset.}
\label{fig:WHAT}
\end{figure}

The spread of news and articles is crucial for shaping public understanding and awareness of climate change and environmental pollution. Media coverage plays a vital role in informing the public, driving policy decisions, and mobilizing climate action \cite{HASE2021102353,Olausson2009,Nassanga2016}. However, this vital communication process faces significant challenges due to the widespread dissemination of misinformation and disinformation. These issues are exacerbated by the overwhelming amount of content generated online, leading to information overload. In this saturated environment, manipulative and misleading content often circulates unchecked, creating confusion in public debates and distorting the science behind climate change \citep{HASE2021102353}. This confusion contributes to political inaction, as conflicting narratives slow down decision-making processes, dilute the urgency of the issue, and hinder the implementation of effective climate policies \citep{Benegal2018,Ding2011}. Misinformation, whether intentional or accidental, has the power to shape public perceptions, leading to skepticism or denial of climate change, ultimately weakening collective action efforts \citep{Olausson2009}.

In this context, Natural Language Processing (NLP) emerges as a critical tool for addressing these challenges. By enabling the automated processing, analysis, and classification of vast amounts of textual data, NLP can help identify, filter, and correct misleading information while amplifying scientifically accurate content. This becomes especially important in low-resource language regions, such as Bengali (Bangla)-speaking communities, where access to accurate climate information is limited. Through NLP, improved information dissemination can reach the populations most affected by climate change, ensuring that they have the knowledge necessary to advocate for and implement more effective climate policies \citep{stede-patz-2021-climate,Nicholls2020}. NLP can also enhance the accessibility of global climate discussions, helping to break down language barriers that prevent these communities from engaging in critical debates and receiving the latest scientific developments \citep{Nicholls2020}.

Low-income countries often face the dual challenges of lacking the technological and linguistic resources to engage fully in global discussions on climate change, even though they are among the most vulnerable to its impacts \citep{Nassanga2016,Olausson2009}. These countries predominantly use languages other than English, which limits their ability to communicate their concerns on the international stage. Despite being home to millions of people directly affected by environmental changes, their voices remain marginalized in global forums. For example, while Bengali (Bangla) is spoken by over 300 million people worldwide, research related to climate change and NLP in this language remains severely underdeveloped \citep{SDCG,Naddaf2022}. This lack of focus on Bengali-language datasets, machine learning models, and research not only hinders effective climate communication but also perpetuates the global knowledge gap regarding the specific climate challenges faced by Bangladesh, a country frequently cited as one of the most at-risk due to climate change \citep{GlobalClimateRiskIndex2021}.

To overcome these issues, we present \texttt{Dhoroni}, a novel multi perspective Bangla (Bangali0 climate change and environmental news dataset. \texttt{Dhoroni} (pronounced "Dharaṇī") signifies "mother earth" in Bengali (Bangla), inspiring us to preserve our birthplace, Mother Earth, and protect the environment to prevent climate change. In addition to the dataset, we introduce a fmaily of baseline Bengali (Bangla) language models, named \textbf{\texttt{BanglaBERT-Dhoroni}}, fine-tuned on this dataset. 
This research gap creates a significant barrier to the effective processing and sharing of climate-related information from Bangladesh within the global scientific community. By underutilizing the potential of NLP for languages like Bengali (Bangla), the international conversation surrounding climate change risks becoming skewed, favoring voices from more developed, English-speaking nations. This further delays the recognition and action needed to address the environmental crises faced by nations like Bangladesh. Therefore, it is imperative to prioritize developing NLP technologies for low-resource languages to ensure a more equitable and inclusive approach to global climate solutions.

Our core contributions are summarized as below:
\begin{enumerate}

    \item We introduce \texttt{Dhoroni}, a novel multi-perspective benchmark dataset, containing a collection of 2300 Bengali (Bangla) news articles from various sources and perspectives worldwide.
    \item  We annotate these news articles by three different annotators, based on ten different perspectives, including (i) stance detection, (ii) political influence, (iii) scientific or (iv) statistical data mentioned, (v) data source or report references, (vi) authenticity, (vi) location impacted, (vii) climate/environmental topic focused on the news, (ix) target of news, and (x) authority involvement.
    \item We also provide a detailed exploratory analysis of the dataset. We define and thoroughly discuss the reasoning behind each perspective, as well as the various annotation options. This is accompanied by a comprehensive exploratory statistical analysis to offer deeper insights into the dataset.
    \item We introduce a series of 10 new baseline models for identifying under \textbf{\texttt{BanglaBERT-Dhoroni}} family, one for each perspective discussed in the \texttt{Dhoroni} dataset in Bengali. These models are developed by fine-tuning \texttt{BanglaBERT} on the \texttt{Dhoroni} dataset, accompanied by detailed performance analytics.
\end{enumerate}

Our \texttt{Dhoroni} dataset and \texttt{BanglaBERT-Dhoroni} model family leverage natural language processing to analyze Bengali climate discourse, providing insights into environmental management, policy, and economics. This initiative aids in the development of sustainable policies, economic valuation of resources, and socio-economic impact analysis of climate change. It informs decision-making on waste management, pollution control, and environmental justice, while fostering public engagement and pro-environmental behavior. Ultimately, it supports policy effectiveness and bridges gaps between policy-makers and local communities.

In the following sections, we will cover the following topics: related works in Section \ref{sec:RelatedWorks}, \texttt{Dhoroni} dataset development methodologies and considerations in Section \ref{sec:dataset-dev}, exploratory analysis of the dataset in Section \ref{sec:EDA-dataset}, \texttt{BanglaBERT-Dhoroni} model family design and development in Section \ref{sec:model-experiments} and benchmarking experiments with experimental findings and analysis in Section \ref{sec:benchmarking-experiments}, and discussions in Section \ref{sec: discussions}.

\section{Related Works} \label{sec:RelatedWorks}
The intersection of climate change and NLP has become increasingly important, with researchers exploring how NLP can contribute to understanding and addressing climate challenges. \cite{stede-patz-2021-climate, 10.1145/3485128} emphasize the significance of integrating NLP with social science to navigate complex climate discussions, providing valuable insights for policy-making and activism. \cite{mallick2024analyzing} highlight the role of NLP in understanding regional climate impacts, stressing the importance of localized adaptation and mitigation efforts. Furthermore, studies on climate awareness in NLP research emphasize the importance of fostering climate-conscious approaches within the NLP community, advocating for technology-driven solutions that contribute to global climate action \citep{hershcovich-etal-2022-towards}. 

In contrast, existing research in NLP applications related to climate change is limited, particularly in languages like Bengali (Bangla). \cite{ni-etal-2023-chatreport} introduce ChatReport, facilitating automated analysis of corporate sustainability reports and addressing challenges like LLM hallucination. \cite{luo-etal-2020-detecting} contribute DeSMOG, a dataset focusing on opinion-framing in the global warming debate, while \cite{Codsgsan2021} develops a taxonomy of climate contrarianism and a computational model for claim classification, revealing insights into contrarian discourse evolution. Additionally, CLIMATE-FEVER \citep{DBLP:journals/corr/abs-2012-00614} improves algorithms for verifying climate claims, and CLIMATEBERT \citep{webersinke2022climatebert} proposes a transformer-based model pretrained on climate texts, enhancing classification and sentiment analysis. While these studies demonstrate diverse research efforts, ranging from automated report analysis to understanding public opinion dynamics and combating climate change misinformation, they are primarily focused on English-language datasets and models. Consequently, they do not adequately represent the perspectives and challenges faced by many other low-resource, low-income, highly climate change-affected countries worldwide.

Inspired by these recent studies and the research gap, our work introduces a novel dataset of Bengali (Bangla) news articles, enabling extensive climate and environmental discourse analysis using NLP, covering various perspectives such as stance detection, political influence, authenticity, impacted locations, target, and authority involvement. Focused on Bengali, spoken in a leading climate change-affected region, our work amplifies underrepresented voices in global climate discussions. This dataset offers valuable insights for advancing NLP applications in climate discourse, addressing specific challenges of climate communication in Bengali (Bangla)-speaking regions.

\begin{figure}[t] 
\centering {\includegraphics[width=\textwidth]{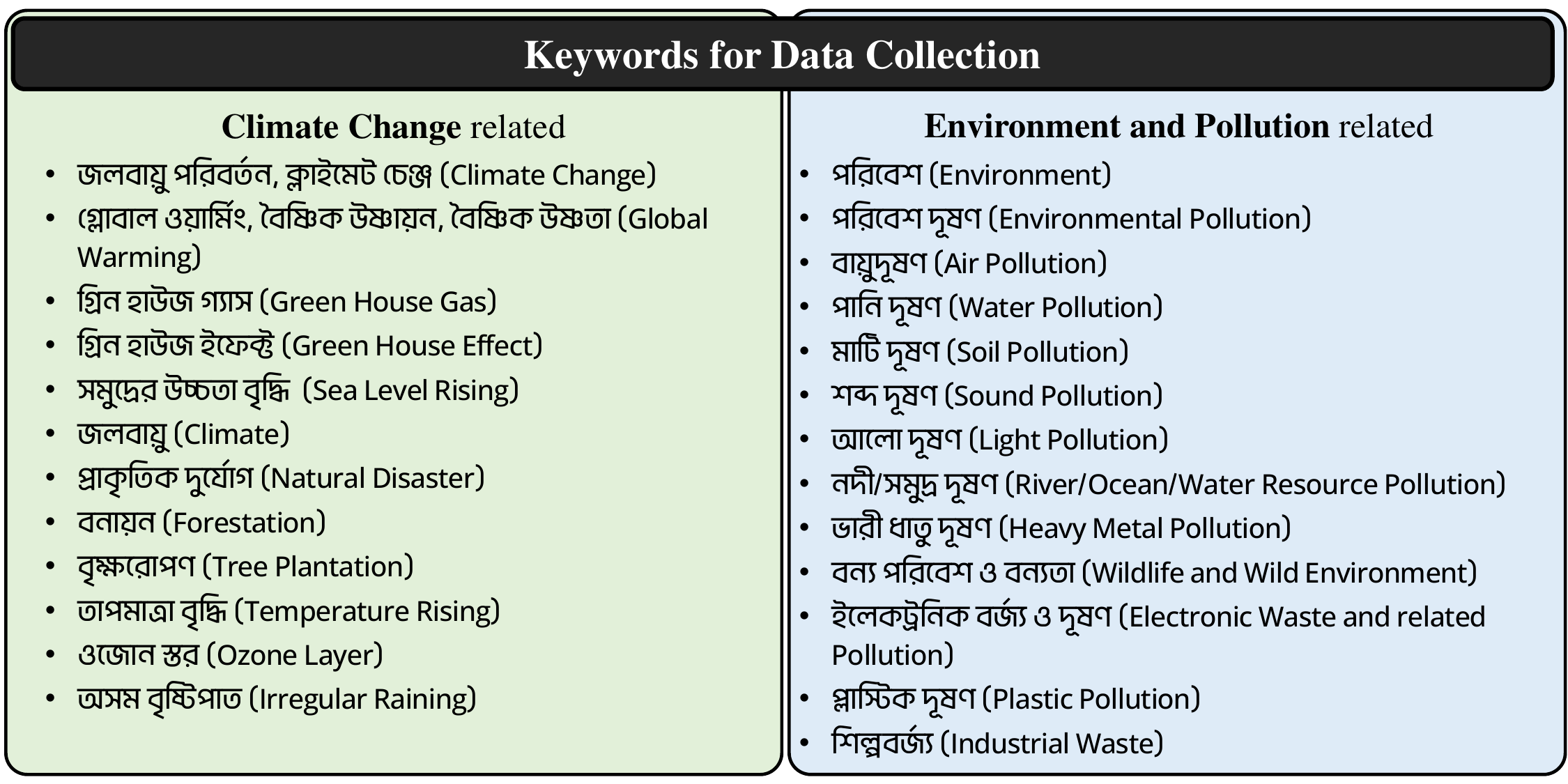}}
\caption{Keyword used for data collection in \texttt{Dhoroni} dataset.}
\label{fig:keywords}
\end{figure}

\section{Methodology: \textit{Dhoroni} Dataset} \label{sec:dataset-dev}
We divide the dataset development methodology into three core parts: (i) Raw data collection, (ii) Data parsing, and (iii) Data curation and annotation. Each of these steps is discussed in detail below.

\subsection{Raw Data Collection} \label{sec:data-collection}
\textbf{Keywords Selection.}
To ensure comprehensive coverage of climate change and environmental pollution topics in our dataset, we meticulously identified and selected a wide range of keywords that reflect the most critical and relevant concepts within these domains. The selection process was informed by both the scientific literature and commonly used terms in public discourse related to climate and environmental issues. Below, we provide an explanation for each category of keywords. Bengali (Bangla) translation of the keywords are included in Figure \ref{fig:keywords}.

\begin{enumerate}
    \item  \textbf{Climate Change-Related Keywords:} 

\begin{itemize} 
\item \textit{Climate Change}: A fundamental term directly related to the overarching phenomenon of long-term alterations in weather patterns, driven largely by human activities. 
\item \textit{Global Warming}: Focusing on the rise in Earth's average temperature, this keyword captures a key driver of climate change. 
\item \textit{Greenhouse Gas} and \textit{Greenhouse Effect}: These terms are essential for capturing the role of gases like carbon dioxide and methane, which trap heat in the atmosphere. 
\item \textit{Sea Level Rising}: One of the most significant impacts of climate change, especially for low-lying regions. 
\item \textit{Climate}: A broad term that captures discussions about climate variability, patterns, and long-term changes. 
\item \textit{Natural Disaster}: Captures the increasing frequency of disasters like floods, droughts, and hurricanes, linked to climate change. 
\item \textit{Forestation} and \textit{Tree Plantation}: These keywords are related to mitigation strategies such as increasing forest cover to absorb carbon dioxide. 
\item \textit{Temperature Rising}: Refers to the increase in global and regional temperatures, a key climate change effect. 
\item \textit{Ozone Layer}: While not directly related to global warming, this term is important for environmental protection efforts. 
\item \textit{Irregular Raining}: Highlights erratic rainfall patterns caused by climate change, affecting agriculture and ecosystems.
\end{itemize}

\item \textbf{Environment and Pollution-Related Keywords:}
\begin{itemize}
    \item \textit{Environment}: A fundamental term used to frame discussions on environmental issues and conservation efforts.
    \item \textit{Environmental Pollution}: A broad term that encompasses all forms of pollution degrading natural ecosystems.
    \item \textit{Air Pollution}: One of the most visible forms of pollution, contributing to health problems and exacerbating climate change.
    \item \textit{Water Pollution}: Refers to contamination of freshwater and marine ecosystems, crucial for human and environmental health.
    \item \textit{Soil Pollution}: Closely related to industrial and agricultural activities, with long-term impacts on food security and ecosystems.
    \item \textit{Sound Pollution}: An important environmental issue, especially in urban settings, affecting both humans and wildlife.
    \item \textit{Light Pollution}: Emerging as a significant issue that disrupts ecosystems, particularly nocturnal wildlife.
    \item \textit{River/Ocean/Water Resource Pollution}: Water bodies are vulnerable to pollution from industrial, agricultural, and residential waste.
    \item \textit{Heavy Metal Pollution}: Heavy metal contamination from industrial activities poses serious health and environmental risks.
    \item \textit{Wildlife and Wild Environment}: Ensures that the dataset captures the effects of environmental degradation on biodiversity and ecosystems.
    \item \textit{Electronic Waste and related Pollution}: A growing issue with the rise of technology, contributing to toxic pollution in landfills.
    \item \textit{Plastic Pollution}: A global environmental crisis impacting oceans, wildlife, and human health.
    \item \textit{Industrial Waste}: Captures pollution from industrial processes, which is a major source of environmental degradation.
\end{itemize}
\end{enumerate} 

\begin{figure*}[t] 
\centering {\includegraphics[width=\textwidth]{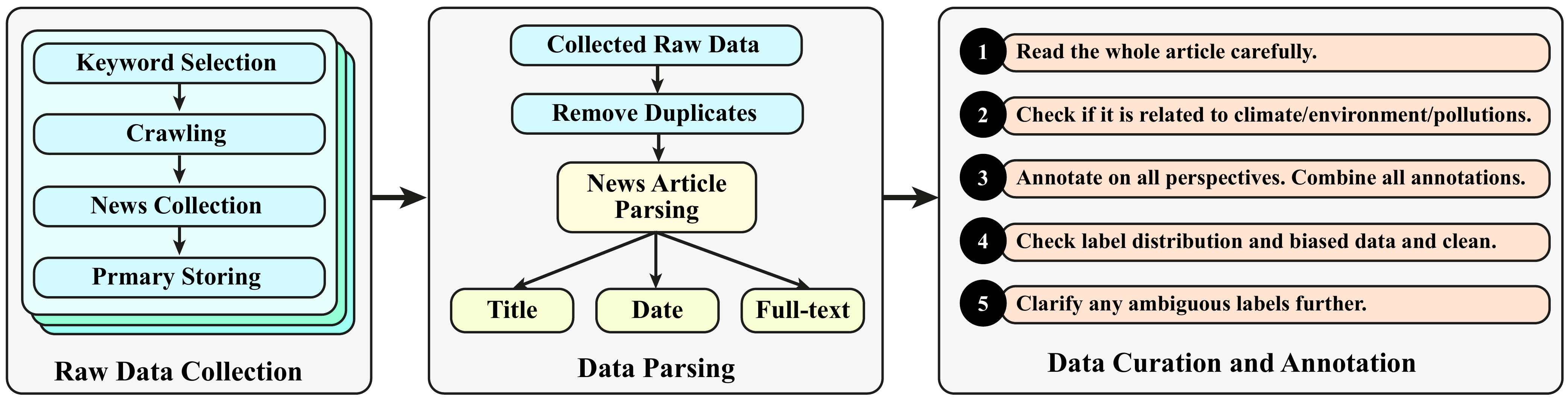}}
\caption{Overview of Data Collection Methodology.}
\label{fig:DCM}
\end{figure*}

\textbf{Crawling, News Collection and Primary Storing.}
To ensure the dataset captures a wide spectrum of climate change and environmental topics, we aim to collect a minimum of 300 articles per keyword. However, the availability of articles can vary significantly depending on the keyword, with more prevalent terms like \textit{Climate Change} and \textit{Air Pollution} having greater media coverage, while terms like \textit{Electronic Waste} or \textit{Heavy Metal Pollution} might be less frequently discussed. 
To address this, we repeatedly source articles from a variety of news platforms, blogs, and other relevant websites. Despite our thorough approach and continuous efforts, some scarcity in specific topics is still observed, particularly for niche areas or technical news. We recognize that this is expected in cases where certain environmental issues receive less media attention or public discourse. Nevertheless, by diversifying our sources and leveraging multiple types of content, we aim to mitigate these gaps and ensure a well-rounded dataset that still reflects the breadth of climate change and environmental pollution topics. Once collected, the data is consolidated and standardized for consistent analysis and storage, allowing for efficient processing and retrieval during model training and testing phases.

\subsection{Data Parsing}
After gathering the raw data, our next step in the data processing pipeline is to remove duplicates before parsing the news articles. Removing duplicates is essential to maintain the integrity and uniqueness of our dataset. Duplicates can occur due to technical issues during data collection or from crawling the same webpages multiple times. Additionally, articles that contain multiple keywords may appear as duplicates and need to be identified and eliminated. Once duplicates are removed, we move on to the parsing stage where we structure the extracted data into key components like title, date, and body text. This process ensures that each article's important information is well-organized and ready for further analysis.

\subsection{Data Curation and Annotation}
\textbf{Data Curation.}
To ensure a focused dataset, we implemented a systematic approach to select relevant articles. First, we thoroughly read each article to understand its content and context. Next, we identified whether the article was related to climate change, the environment, or pollution, with a focus on these specific themes. Initially, we collected approximately 2,830 raw articles, but after processing and annotation, only 2,300 articles were deemed relevant to our objectives.

\textbf{Exploring Different Perspectives.}
We then proceeded to annotate each article based on the perspectives outlined in Figure \ref{fig:WHAT}, capturing key insights and viewpoints presented in the text. Details on each perspective, including their definitions, meanings, and related statistics, are provided in Section \ref{sec:perspectives}. To ensure the quality and objectivity of our dataset, we examined the distribution of labels to identify any potential biases. If biases were detected, we took corrective measures to clean the data and ensure a balanced representation of perspectives. Finally, we thoroughly reviewed the annotations to resolve any ambiguous labels, ensuring clarity and accuracy throughout the data annotation process.

\textbf{Annotation.}
Each sample was independently annotated by three annotators, and their results were consolidated into a single final annotation. Any conflicts in labelling were resolved collaboratively by the authors, who carefully reviewed the differing perspectives and determined the most appropriate label for each piece of news. Additionally, we developed custom annotation software using Python, which is available at project GitHub repository : \textbf{\textit{\href{}{}}}.

\subsection{Exploratory Data Analysis: \texttt{Dhoroni} Dataset}  \label{sec:EDA-dataset}
In this section, we present an exploratory data analysis of the \texttt{Dhoroni} dataset. We examine the overall structure and content of the dataset, identifying key trends, patterns, and distributions. This analysis provides a comprehensive understanding of the dataset's composition, aiding in the interpretation of subsequent results.

\subsection{Dataset Format and Statistics}
The dataset consists of 2,300 properly annotated articles. It includes news from 2022 to 2024, collected between April and August 2024. Since most Bengali (Bangla) speakers reside in Bangladesh and West Bengal, India, the news articles were sourced from newspapers covering these regions. While the primary language is Bengali, occasional English or Hindi words appear, though minimally. The annotation data is available in table format, with full news texts provided in \texttt{.txt} files.

\begin{figure}[t]
    \centering
    \includegraphics[width=\linewidth]{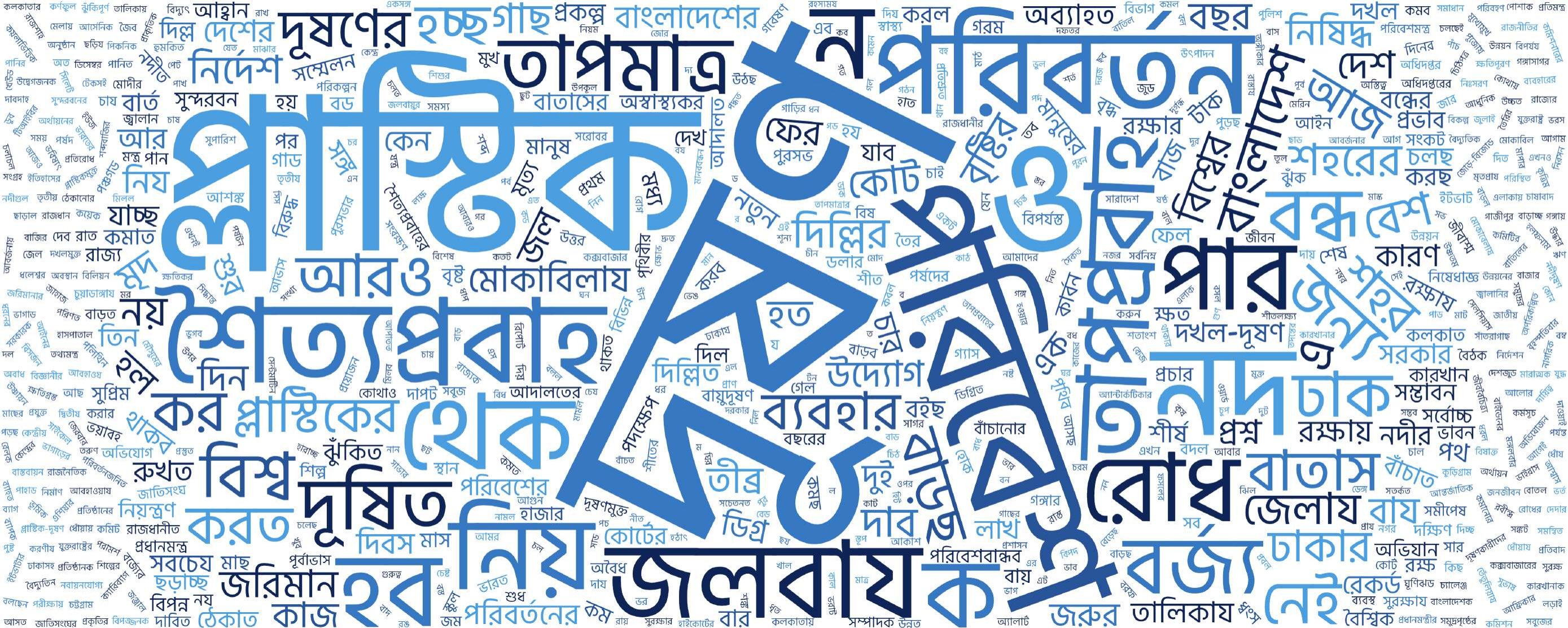}
    \caption{Word cloud of news titles.}
    \label{fig:sc-news-titles}
\end{figure}

\begin{figure}[t]
    \centering
    \includegraphics[width=\linewidth]{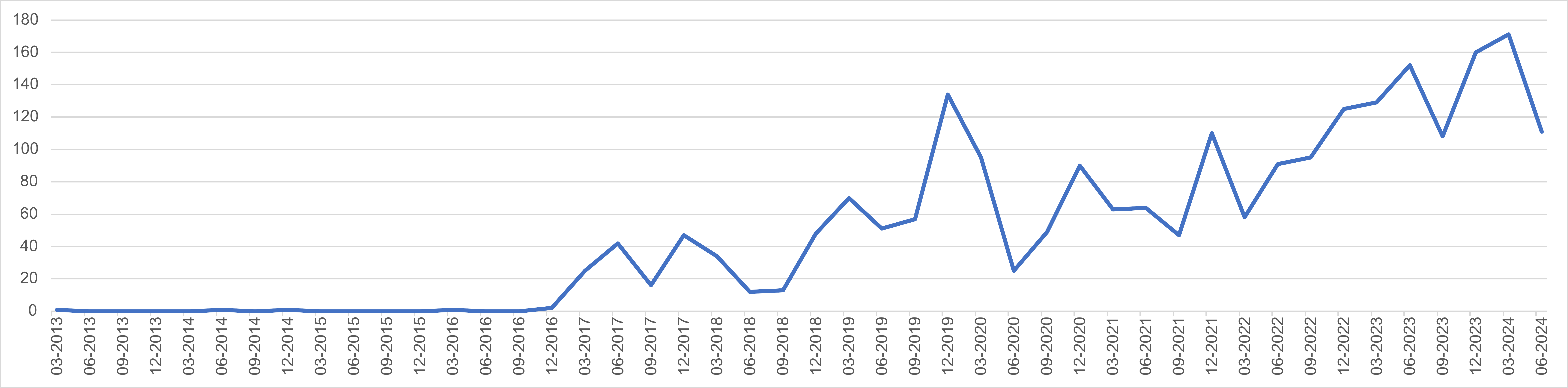}
    \caption{Number of News Articles Collected from Each Quarter.}
    \label{fig:sc-news-timeline}
\end{figure}

The titles in the dataset have an average of about 6.65 words, with each title containing approximately 45.21 characters. This gives an average word length of 6.06 characters for the titles. On the other hand, the text sections have an average word count of 394.24, consisting of around 2615.22 characters. The average word length in the text is about 5.66 characters. Each text section is made up of an average of 29.56 sentences. In these sentences, there are on average 14.79 words and 98.78 characters per sentence. 
Figure \ref{fig:sc-news-titles} shows a word cloud of the news titles. The terms \textit{Pollution, Environment, Plastic, Climate, Temperature, Change, Heatwave, Plastic} and \textit{River}\footnote{English translation of the words are added here for more clarity and understanding.} are the most visible ones. This word cloud aligns perfectly with our keyword design and data collection methodology.
Figure \ref{fig:sc-news-timeline} shows how many news articles are collected from each quarter from 2013 to 2024. In the Figure, it is also evident that the number of news articles began to increase significantly starting in the fourth quarter of 2016.

\subsection{Definition and Statistics of Each Perspective} \label{sec:perspectives}
In this section, we discuss definitions and statistics associated with the various perspectives captured within the \texttt{Dhoroni} dataset. We analyze the distribution of perspectives, examining how each viewpoint is represented across the dataset. This analysis sheds light on the diversity and balance of perspectives, offering insights into the dataset's coverage and potential biases.

\begin{figure}[t]
\centering
\subfigure[Stance Detection]{%
\resizebox*{6cm}{!}{\includegraphics[scale=0.6]{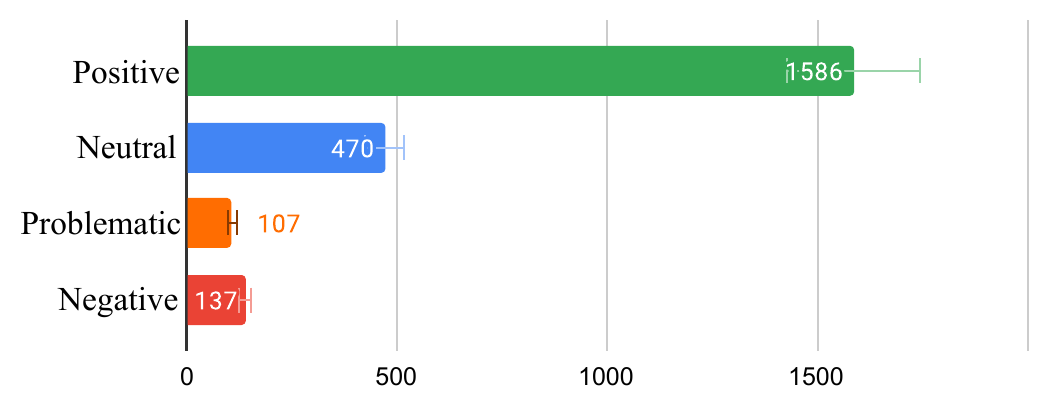}}}
\subfigure[News Authenticity Identification]{%
\resizebox*{6cm}{!}{\includegraphics[scale=0.6]{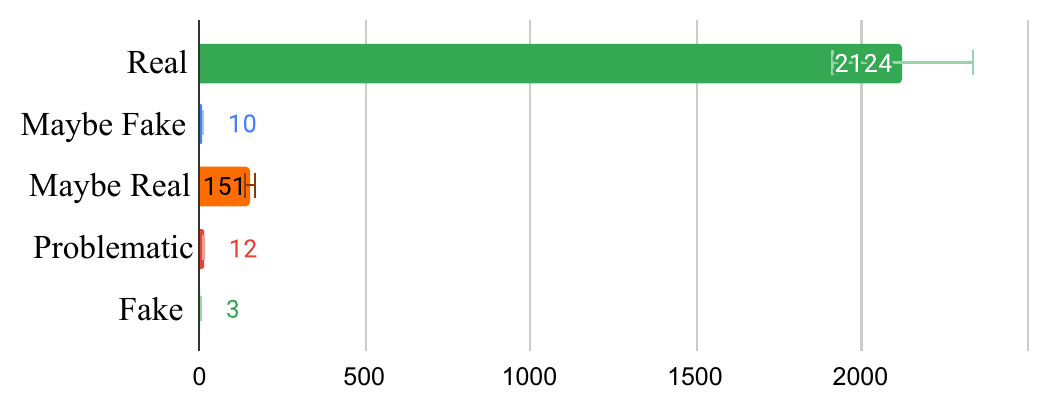}}}

\caption{Label Distribution of \textit{Stance Detection and News Authenticity Identification} in \texttt{Dhoroni}.} \label{fig:EDA-6}
\end{figure}

\subsubsection{Stance Detection}
Stance detection in this context refers to identifying whether a news article supports or opposes the concept of climate change, or remains neutral or unclear. The options available are \textit{Positive, Negative, Neutral, and Problematic}. A Positive stance indicates that the news supports climate change as a real and pressing issue, for example, an article discussing the impact of rising sea levels due to global warming. A Negative stance refers to news that opposes the idea of climate change, often claiming it as fake or propaganda. For instance, a report suggesting that climate change is a hoax would be labeled as Negative. Neutral means the article neither supports nor opposes climate change, staying objective without clear bias. Finally, Problematic means the stance is unclear or fluctuates within the article, potentially presenting mixed viewpoints that make it difficult to categorize. An article that first acknowledges climate change but later questions its validity could be deemed Problematic.

In Figure \ref{fig:EDA-6}(a), we can see the label distribution for stance detection. We have 2,300 total instances, with the majority being positive (1,586) and neutral (470) stances. Negative stances accounted for 137 instances, while 107 were identified as problematic. This distribution shows a stronger presence of positive and neutral perspectives within the data.

\subsubsection{News Authenticity Identification}
News authenticity identification is the process of determining the trustworthiness of a news article. The options are \textit{Fake, Real, Maybe Fake, Maybe Real, and Problematic}. An article confirmed to spread false information or conspiracies would be labeled Fake, while a credible article based on verified facts would be considered Real. Maybe Fake refers to articles that could potentially contain misinformation but lack definitive evidence, while Maybe Real suggests the news seems reliable but may not have enough confirmation. Problematic is used when the article's authenticity is difficult to assess, such as when it mixes real facts with potential misinformation, making it unclear.

In Figure \ref{fig:EDA-6}(b), we can see the label distribution for news authenticity detection. The majority of the news items were classified as real (2,124), with a smaller portion labeled as maybe real (151). Instances marked as fake were minimal (3), while 10 were categorized as maybe fake, and 12 were identified as problematic. This distribution suggests a high proportion of authentic news in the dataset.

\subsubsection{Political Influence Detection}
Political influence detection involves identifying whether the news article shows signs of political motives or influence concerning climate change. The options for this detection are \textit{Yes or No}. If the article appears to support or oppose climate change due to political biases, for instance, if it promotes a government policy that denies climate change for political gains, the answer would be 'Yes'. On the other hand, if the article remains purely factual or independent of political agendas, it would be labeled as 'No'.

\begin{figure}[t]
\centering
\subfigure[Political Influence Detection]{%
\resizebox*{6cm}{!}{\includegraphics[scale=0.6]{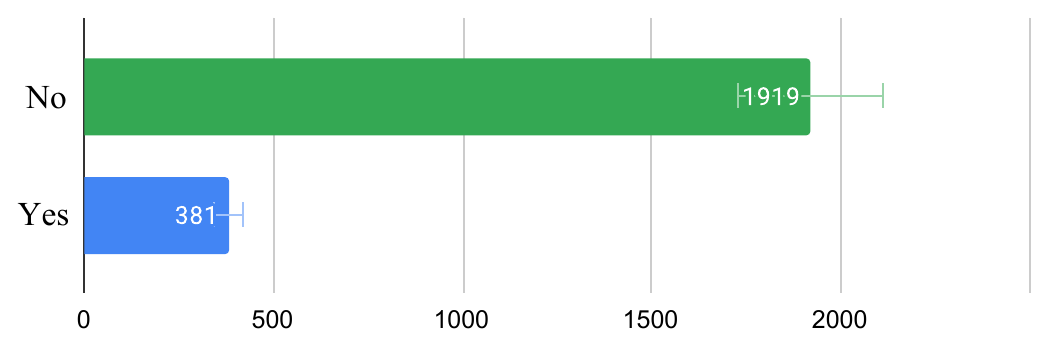}}}
\subfigure[Scientific Data Usage Detection]{%
\resizebox*{6cm}{!}{\includegraphics[scale=0.6]{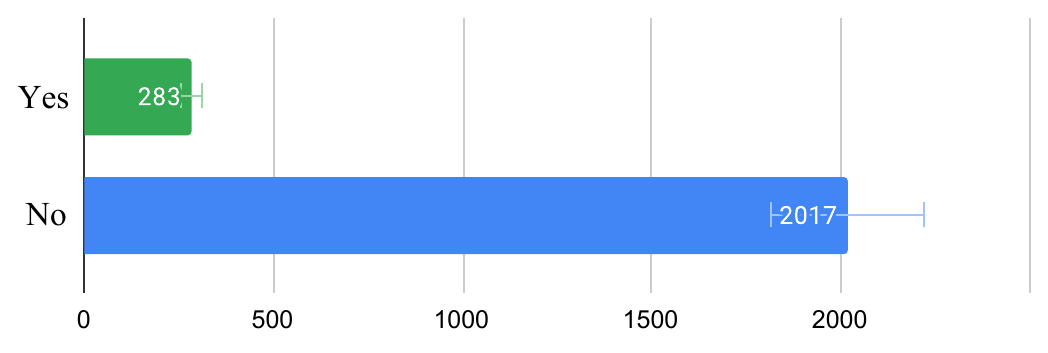}}}

\subfigure[Statistical Data Usage Detection]{%
\resizebox*{6cm}{!}{\includegraphics[scale=0.6]{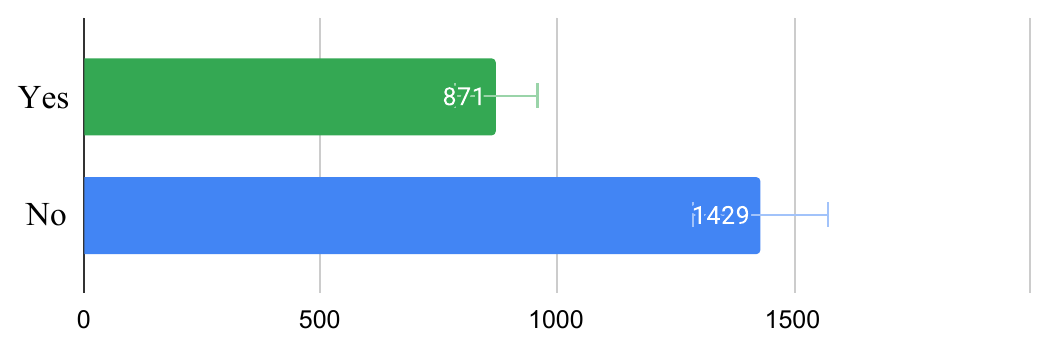}}}
\subfigure[Data Sources Detection]{%
\resizebox*{6cm}{!}{\includegraphics[scale=0.6]{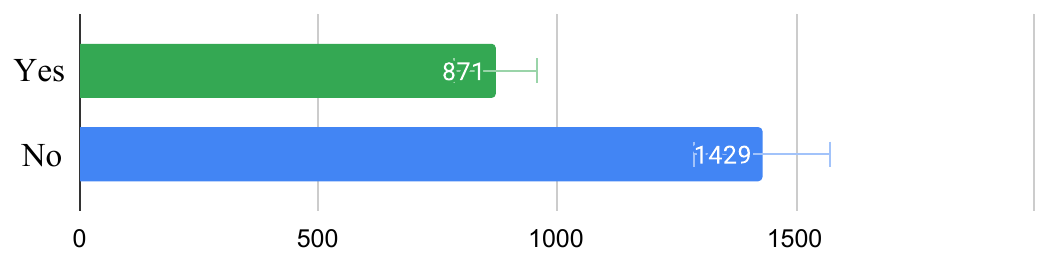}}}

\caption{Label Distribution of \textit{Political, Scientific and Statistical Influence and Data Sources} in \texttt{Dhoroni}.} \label{fig:EDA-7}
\end{figure}

In Figure \ref{fig:EDA-7}(a), it shows that 381 instances were marked as politically influenced, while the majority, 1,919, had no political influence. This indicates there are some politically influenced news, but not that much.

\subsubsection{Scientific Data Usage Detection}
Scientific data usage detection refers to checking whether the news article includes or cites scientific data to support its claims. The options are \textit{Yes or No}. If the article references studies, scientific findings, or expert opinions about climate change, the answer would be Yes. For example, an article discussing the effects of ${CO}_2$ emissions with references to scientific research would be marked as 'Yes'. If no scientific data is used, the answer would be 'No'.

In Figure \ref{fig:EDA-7}(b), it shows that most news items (2,017) did not mention any scientific data, with only 283 including such references. The lack of scientific data in the majority of reports raises potential concerns about their authenticity.

\subsubsection{Statistical Data Usage Detection}
Statistical data usage detection examines whether the article employs any statistical data to validate its arguments. The options here are \textit{Yes or No}. If the article includes numbers, percentages, or trends, for example, showing the rise in global temperatures by $1.5\circ C$ over the past century, it would be labelled 'Yes'. If no statistics are used, the answer would be 'No'

In Figure \ref{fig:EDA-7}(c), it shows that most news items (1,429) did not mention any statistical data, while 871 did. The absence of statistical data in many reports may raise concerns about their reliability.

\subsubsection{Data Sources Detection}
Data sources detection determines whether the article mentions the origin of any scientific or statistical data cited within it. The options are \textit{Yes or No}. If an article provides citations, such as referencing the Intergovernmental Panel on Climate Change (IPCC) or a specific research institution, the answer would be 'Yes'. If the article presents data without mentioning its source, the answer would be 'No'.

In Figure \ref{fig:EDA-7}(d), it shows that most news items (1,672) did not mention any sources, with only 628 providing them. This lack of source attribution in the majority of reports may raise questions about their credibility.

\subsubsection{Impacted Location or Countries Detection}
Impacted location or countries detection identifies whether the news specifies particular locations or countries affected by climate change. The options are the name of a country or global. For example, if an article discusses how Bangladesh is affected by rising sea levels, it would be tagged with Bangladesh. If the article addresses global impacts, such as rising global temperatures or widespread environmental effects, it would be labeled global.

\begin{figure}[h]
    \includegraphics[width=\linewidth]{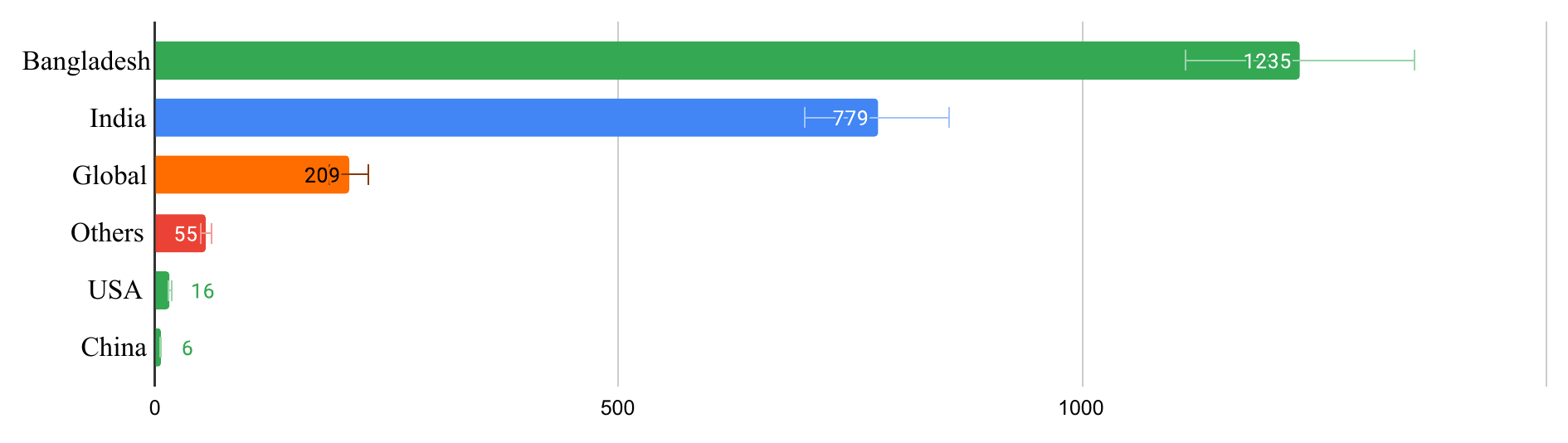}
    \caption{Label Distribution of \textit{Impacted Location or Countries Detection} in \texttt{Dhoroni}.}
    \label{fig:ImpactedLocationOrCountriesDetection}
\end{figure}

In Figure \ref{fig:ImpactedLocationOrCountriesDetection}, we can see the label distribution for impacted location of countries. The dataset primarily focuses on Bangladesh, with 1,235 entries, and West Bengal, India, with 779 entries, reflecting its regional emphasis. Global entries account for 209 instances, while other countries like the USA (16) and China (6) are also represented, indicating the presence of major global powers. The smaller representation of these countries is consistent with the dataset's primary focus, yet their inclusion highlights the broader relevance of global affairs within the content. Other locations are less frequent, with only 55 entries, mostly consisting of 1-2 mentions of individual countries. We grouped these under "Others" to make the label distribution more balanced.

\subsubsection{Detection of Climate/Environmental Topics}
Climate/environmental topic detection checks whether the news article covers subjects related to climate change or environmental issues. No specific options are provided, as the detection simply confirms the presence of such topics. For instance, an article discussing deforestation, carbon emissions, or renewable energy would fall under this category. Most of the options here are from keywords, described in Section \ref{sec:data-collection}.

\begin{figure}[h]
    \includegraphics[width=\linewidth]{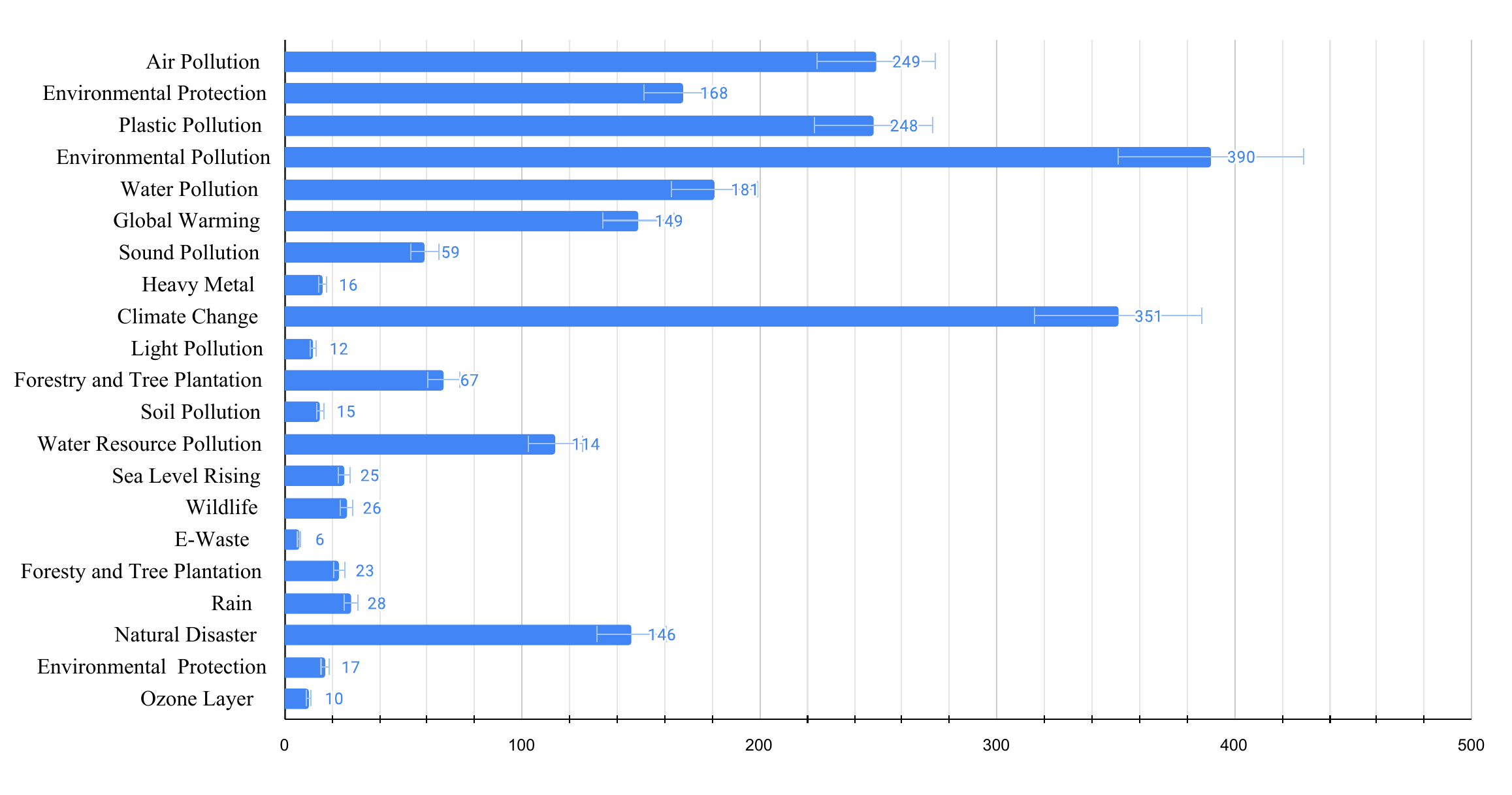}
    \caption{Label Distribution of \textit{Detection of Climate/Environmental Topics} in \texttt{Dhoroni}.}
    \label{fig:DetectionofClimate_EnvironmentalTopic}
\end{figure}

In Figure \ref{fig:DetectionofClimate_EnvironmentalTopic}, we can see the label distribution for various topics related to climate change and the environment. The most prominent categories include Environmental Pollution (390), Climate Change (351), and Air Pollution (249), indicating a strong focus on these areas. Plastic Pollution also appears frequently, with 248 entries, highlighting concerns about waste management. Water-related issues, such as Water Pollution (181) and Water Resource Pollution (114), are also well-represented. In contrast, topics like E-Waste (6) and Ozone Layer (10) have fewer mentions, suggesting these issues receive less attention in the dataset. Overall, the focus is heavily on pollution and climate change, which aligns with global environmental concerns.

\subsubsection{News Target Detection}
News target detection identifies the groups or entities blamed or held responsible in the news article regarding climate issues. The options are \textit{N/A, All, Government, General People, Industries, Dishonest People, Authority, or World Power}. For instance, if an article blames industrial corporations for their carbon emissions, it would be labeled Industries. If the article suggests that everyone shares the responsibility, it would be tagged All. Dishonest People would apply when the blame is placed on individuals who mislead or manipulate climate-related information. Authority refers to governing or regulatory bodies, and World Power points to influential global actors like the UN or major countries.

\begin{figure}[h]
    \includegraphics[width=\linewidth]{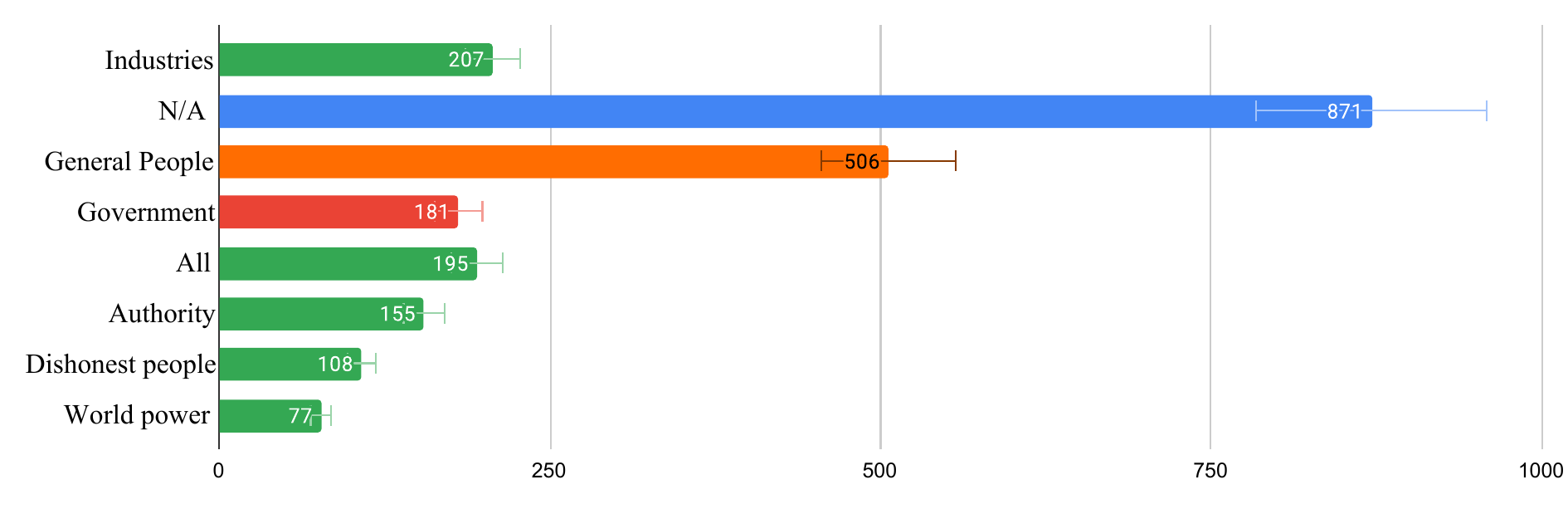}
    \caption{Label Distribution of \textit{News Target Detection} in \texttt{Dhoroni}.}
    \label{fig:NewsTargetDetection}
\end{figure}

In Figure \ref{fig:NewsTargetDetection}, we can see the distribution of targets. General People are the most frequently mentioned group, with 506 instances, indicating that many articles hold the public accountable for climate-related issues. Industries also appear prominently with 207 mentions, reflecting concerns about corporate environmental practices. Authorities and Government together account for 336 mentions, highlighting their role in policy-making and enforcement. Dishonest people, likely referring to individuals or groups involved in harmful practices, are identified 108 times. World powers are held responsible in 77 cases, underscoring their influence in global climate decisions. A significant portion of the dataset (871) is categorized as N/A, where no specific target is mentioned, indicating that many articles discuss environmental issues without attributing blame to a particular group. This distribution reveals a mix of public, corporate, and governmental responsibility in climate reporting.

\subsubsection{Authority Involvement Detection}
Authority involvement detection checks whether any authority is mentioned in the article and, if so, identifies which one. The options include \textit{N/A, Legal/Court, Crime/Police, Other Authority, Environmental/Forestry Authority, Scientists and Researchers, NGO, Government, General People, or United Nations and such Global Institutions}. For example, if the news discusses a court ruling on environmental regulations, the tag would be Legal/Court. If it mentions police involvement in climate protests, it would be labeled Crime/Police. News referencing global authorities like the UN would be tagged as UN, while articles mentioning scientific research bodies would use Scientists and Researchers.

\begin{figure}[h]
    \includegraphics[width=\linewidth]{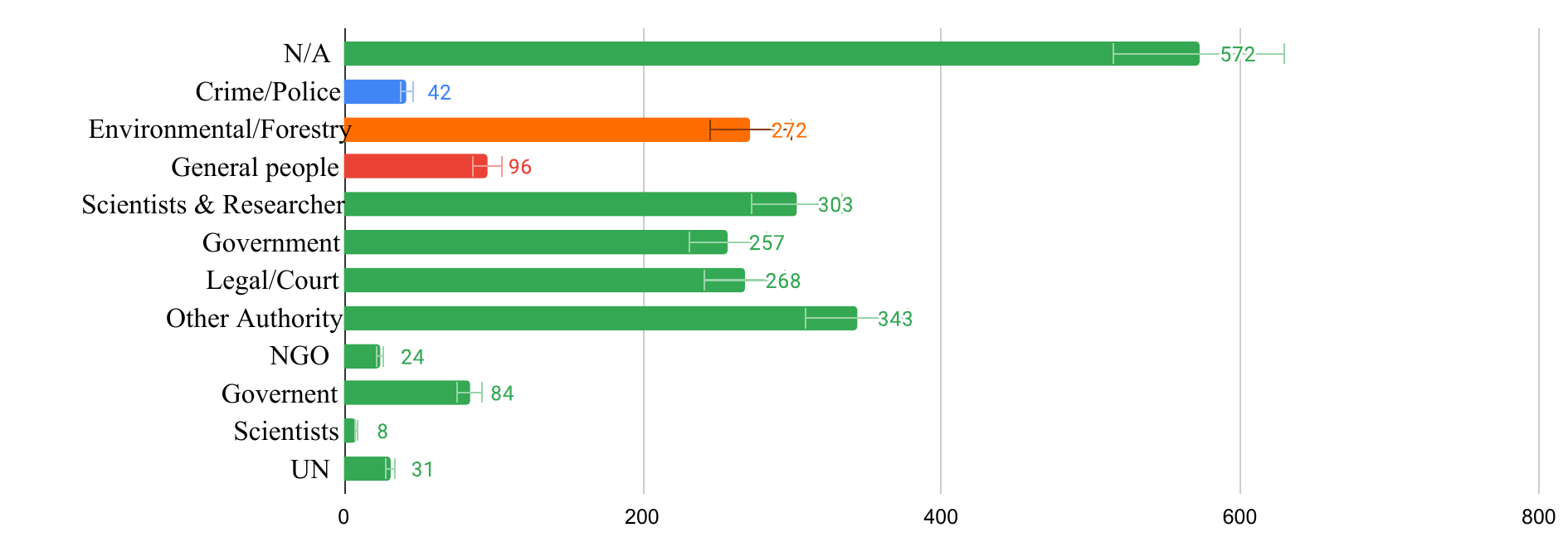}
    \caption{Label Distribution of \textit{Authority Involvement Detection} in \texttt{Dhoroni}.}
    \label{fig:AuthorityInvolvementDetection}
\end{figure}

In Figure \ref{fig:AuthorityInvolvementDetection}, we can see the distribution of authority involvement in the articles. The most frequently mentioned authorities are Environmental/Forestry Authorities (272) and Legal/Court systems (268), reflecting a strong focus on environmental regulation and legal oversight. Scientists and Researchers are cited 303 times, indicating their significant role in climate and environmental discussions. Other authorities, including various unspecified bodies, are mentioned 343 times. Government and Government are noted 341 times collectively, emphasizing the role of governmental bodies in addressing climate issues. Crime/Police authorities are mentioned 42 times, suggesting a smaller focus on enforcement related to environmental crimes. NGOs, UN, and Scientists have fewer mentions, with 24, 31, and 8 instances respectively. A substantial number of articles (572) do not specify any authority, suggesting that many discussions on climate issues do not directly involve particular institutions. This distribution highlights the diverse range of authorities involved in environmental and climate-related reporting, with a notable emphasis on legal, scientific, and regulatory entities.

\subsection{Dataset Potential and Practical Applications} 
We believe, our \texttt{Dhoroni} dataset can serve as a valuable resource for advancing research and applications in the field of natural language processing (NLP) for Bengali. Given the increasing importance of climate change and environmental issues, the dataset provides an extensive collection of 2,300 news articles that cover various perspectives. This breadth of information allows researchers and practitioners to explore a range of applications, from stance detection to the assessment of political influences and the authenticity of sources. Such multifaceted analysis can enable the identification of trends in public discourse, enhancing our understanding of how climate-related topics are framed in the media.

The potential applications of this dataset extend beyond academic research. For instance, it can be utilized by policymakers and NGOs to assess public sentiment regarding environmental issues in Bangladesh. By understanding how the media portrays climate change, these organizations can tailor their communication strategies to engage more effectively with the public. Additionally, businesses focused on sustainability can leverage insights from the dataset to better understand consumer concerns and preferences related to environmental issues. The dataset can inform marketing strategies and corporate social responsibility initiatives, ultimately leading to more environmentally conscious practices.
Moreover, educational institutions can use the \texttt{Dhoroni} dataset as a teaching tool in various disciplines, such as media studies, environmental science, and data analytics. Students can analyze the dataset to develop critical thinking skills and gain practical experience in data-driven research methodologies. By fostering an understanding of how climate change is represented in the media, the dataset can contribute to the next generation of environmentally aware citizens and professionals.

Furthermore, the dataset's focus on multiple perspectives allows for a richer exploration of climate narratives. By analyzing the same events through different lenses, researchers can uncover biases in reporting and examine the motivations behind varying viewpoints. This depth of analysis is crucial for developing necessary discussions about climate change and its impacts. As we continue to confront environmental challenges, the ability to dissect and understand these narratives becomes increasingly vital.

\section{Model Design and Development} \label{sec:model-experiments}

\textbf{Model.} \quad
We have fine-tuned \texttt{BanglaBERT} (\href{https://huggingface.co/sagorsarker/bangla-bert-base}{sagorsarker/bangla-bert-base})\footnote{https://huggingface.co/sagorsarker/bangla-bert-base}\footnote{https://github.com/sagorbrur/bangla-bert}, which is freely available in Hugging Face by \texttt{transformers} library \citep{wolf2020huggingfaces} with MIT Licence. It is a pretrained language model of Bengali language using mask language modeling described in BERT \citep{devlin-etal-2019-bert}.

\textbf{Data.} \quad
We have used stratified sampling to achieve a 70:15:15 split for the train, validation, and test sets, respectively, ensuring that the classes remain balanced. From a total of 2,300 samples in \textit{Dhoroni} dataset, the train set had 1,610 samples, while the validation and test sets had 345 samples each. The same split was used for all tasks.

\textbf{Implementation Details.} \quad
We have fine-tuned this base model for each task, as described in Table \ref{tab:model-family}. 
All of our models are trained for a total of 10 epochs, with early stopping.
As most classes in our data are imbalanced, we have applied distributional alignment in each epoch for a better class distribution, leading to a better and balanced training.
We have used an MLP with two-layer hidden architecture for the classification head, with each layer comprising 128 units. It employs a learning rate of 0.001 to facilitate the optimization process. A dropout rate of 0.7 is incorporated into the model architecture to mitigate overfitting. The training process is conducted using a batch size of 128 to process data during each iteration. Model performance is evaluated based on the F1 Score and Accuracy average, a lot of data is imbalanced. All embeddings utilized in the model are generated from the \texttt{Bangla-BERT} model, with a maximum token length set to 768. These hyperparameters are tuned using a grid-search on \textit{Stance Detection} task, as described in Section \ref{sec:grid-ablation}. We have used Kaggle and Google Colab for model training and experiments. In the metrics, we use Accuracy ($\uparrow$), Precision ($\uparrow$), Recall ($\uparrow$), F1 Score ($\uparrow$) from Scikit-learn\footnote{https://scikit-learn.org/stable/api/sklearn.metrics.html\#module-sklearn.metrics} library \citep{scikit-learn}. In multi-class classification, we used macro precision, macro recall and macro F1 score. These macro versions calculate the value for each class individually and then averages these values, treating all classes equally, regardless of their frequency, which can sometimes lead to low scores, as errors in less frequent classes can significantly impact the overall performance. However, they provide stability and help us evaluate the model effectively, regardless of class imbalance.

\textbf{Distributional Alignment.} \quad
We have designed a distributional alignment function to balance the imbalanced dataset by duplicating samples from minority classes so that each class has the same number of samples as the majority class. It begins by counting the occurrences of each class label, \( n_i \) for class \( c_i \), and identifies the maximum count, \( \texttt{max\_count} = \max(n_1, n_2, \dots, n_k) \). For each data point, the original embedding and label are added to new lists. If a class \( c_i \) has fewer samples than \( \texttt{max\_count} \), the function calculates the number of duplicates needed as \( \texttt{duplicates\_needed} = \left\lfloor{(\texttt{max\_count} - n_i})/{n_i} \right\rfloor \), and replicates the sample accordingly. The result is a balanced dataset with \( \texttt{max\_count} \) samples per class, returned as two lists: one for embeddings and another for labels.

\begin{table}[h]
    \caption{Fine-tuned model details of \texttt{BanglaBERT-Dhoroni} family.}
    \label{tab:model-family}
    \centering
    \begin{tabular}{lc}
    \toprule
    Task  & Location \\
    \midrule
    Stance Detection & 
    \href{https://huggingface.co/ciol-research/BanglaBERT-Dhoroni-SD}{ciol-research/BanglaBERT-Dhoroni-SD} \\
    
    News Authenticity Identification & 
    \href{https://huggingface.co/ciol-research/BanglaBERT-Dhoroni-NAD}{ciol-research/BanglaBERT-Dhoroni-NAD} \\
    
    Political Influence Detection & 
    \href{https://huggingface.co/ciol-research/BanglaBERT-Dhoroni-PAD}{ciol-research/BanglaBERT-Dhoroni-PAD} \\
    
    Scientific Data Usage Detection & 
    \href{https://huggingface.co/ciol-research/BanglaBERT-Dhoroni-SDUD}{ciol-research/BanglaBERT-Dhoroni-SDU} \\
    
    Statistical Data Usage Detection & 
    \href{https://huggingface.co/ciol-research/BanglaBERT-Dhoroni-StDU}{ciol-research/BanglaBERT-Dhoroni-StDU} \\
    
    Data Sources Detection & 
    \href{https://huggingface.co/ciol-research/BanglaBERT-Dhoroni-DSD}{ciol-research/BanglaBERT-Dhoroni-DSD} \\
    
    Impacted Location or Countries Detection & 
    \href{https://huggingface.co/ciol-research/BanglaBERT-Dhoroni-ILD}{ciol-research/BanglaBERT-Dhoroni-ILD} \\
    
    Detecting Climate/Environmental Topics & 
    \href{https://huggingface.co/ciol-research/BanglaBERT-Dhoroni-CET}{ciol-research/BanglaBERT-Dhoroni-CET} \\
    
    News Target Detection & 
    \href{https://huggingface.co/ciol-research/BanglaBERT-Dhoroni-NTD}{ciol-research/BanglaBERT-Dhoroni-NTD} \\
    
    Authority Involvement Detection & 
    \href{https://huggingface.co/ciol-research/BanglaBERT-Dhoroni-AID}{ciol-research/BanglaBERT-Dhoroni-AID} \\

    \bottomrule
    \end{tabular}
\end{table}

\section{Experiments and Benchmarking} \label{sec:benchmarking-experiments}
In this section, we describe the benchmarking process and experiments. We use the model and implementation details defined in Section \ref{sec:model-experiments} and evaluate our \texttt{BanglaBERT-Dhoroni} model family based on defined metrics.

\subsection{Experimental Results and Discussion}
Experimental results provided in Table \ref{tab:Experimental-Results} for the various tasks highlight both strengths and challenges in how well the model performs. Across the tasks, we can see a clear pattern: accuracy tends to be higher, but the more insightful metrics—precision, recall, and F1 score—reveal deeper issues, particularly with identifying specific classes or more complicated information.

\begin{table}[t]
    \centering
        \caption{Experimental Results of All \texttt{BanglaBERT-Dhoroni} model family.}
    \label{tab:Experimental-Results}
    \begin{tabular}{lccccc}
    \toprule
No & Task & Accuracy ($\uparrow$) & Precision ($\uparrow$) & Recall ($\uparrow$)  & F1 ($\uparrow$) \\
\midrule
1 & Stance Detection & 0.635 & 0.426 & 0.295 & 0.308 \\
2 & News Authenticity Identification & 0.911 & 0.184 & 0.198 & 0.191 \\
3 & Political Influence Detection & 0.835 & 0.417 & 0.500 & 0.455 \\
4 & Scientific Data Usage Detection & 0.870 & 0.437 & 0.497 &  0.465 \\
5 & Statistical Data Usage Detection & 0.594 & 0.491 & 0.497 &  0.440 \\
6 & Data Sources Detection & 0.623 & 0.481 & 0.485 &  0.480 \\
7 & Impacted Location Detection & 0.536 & 0.138 & 0.175 &  0.134 \\
8 & Detecting Climate/Env. Topics & 0.052 & 0.015 & 0.076 &  0.017 \\
9 & News Target Detection & 0.380 & 0.048 & 0.125 & 0.069 \\
10 & Authority Involvement Detection & 0.183 & 0.030 & 0.102 & 0.044 \\
\bottomrule
    \end{tabular}
\end{table}

In \textbf{Stance Detection}, the accuracy of 0.635 suggests the model performs reasonably well overall, but the low precision (0.426) and recall (0.295) indicate it struggles to correctly identify stances, especially minority or less obvious ones. Stances, such as "support" or "oppose," can be subtle in language, and the model likely finds it difficult to differentiate between them, leading to more incorrect classifications.

Moving on to \textbf{News Authenticity Identification}, while the accuracy is impressively high at 0.911, precision (0.184) and recall (0.198) are extremely low. This is a classic sign of class imbalance: the model may be biased toward classifying most news as authentic, since it's likely the majority class in the training data. As a result, it rarely identifies false news, which heavily skews its precision and recall. A more balanced dataset or targeted strategies to boost minority class detection could help improve these scores.

In \textbf{Political Influence Detection}, we see a better balance between precision (0.417) and recall (0.500), which is reflected in a decent F1 score of 0.455. This suggests the model is able to identify political influence with moderate success, though it still tends to miss some instances or incorrectly classify others. Political influence might vary significantly depending on context, which could explain why the model sometimes struggles to capture it consistently

\textbf{Scientific Data Usage Detection} performs relatively well, with an accuracy of 0.870 and a balanced F1 score of 0.465. The model seems to manage identifying scientific data usage, likely because this task involves more structured or identifiable cues (like specific terminology or phrasing). However, the model’s slight tendency to overestimate scientific usage (higher recall than precision) may lead it to incorrectly flag some cases.

In contrast, \textbf{Statistical Data Usage Detection} has lower accuracy (0.594), though its precision (0.491) and recall (0.497) are more balanced. This suggests the model can identify statistical data usage somewhat effectively, but the overall complexity of the task—where statistical data may appear in various forms—leads to less consistent results. The slight drop in F1 score (0.440) points to misclassification of some borderline cases.

\textbf{Data Sources Detection} shows moderate performance with an accuracy of 0.623, and relatively balanced precision and recall (both around 0.48). This task may be easier for the model to manage because references to data sources are often explicit, making it easier to pick up on relevant patterns. However, there may still be subtle variations in how sources are mentioned, leading to some misclassifications.

On the other hand, \textbf{Impacted Location or Countries Detection} shows poor performance. With a precision of 0.138 and an F1 score of 0.134, it’s clear the model struggles to consistently identify locations or countries affected by events. This might be due to the broad range of ways in which locations are mentioned in text, from indirect references to unclear geographical descriptions, making it harder for the model to extract this information reliably.

A particularly striking failure is seen in \textbf{Detecting Climate/Environmental Topics}, where the accuracy is just 0.052, and both precision and recall are near zero. This suggests the model is almost completely unable to identify climate-related content, likely due to either insufficient training data or highly ambiguous definitions of what constitutes a "climate" topic. It could also be that the model is not tuned to capture the wide variety of terms and context in which climate issues are discussed.

\textbf{News Target Detection} also shows significant challenges, with low precision (0.048) and a recall of only 0.125. Identifying the target of news content is inherently complex, often involving subtle contextual cues that the model seems unable to consistently capture. This might explain the low F1 score of 0.069, as the model struggles to pinpoint the correct targets amidst varying contexts.

Lastly, \textbf{Authority Involvement Detection} presents similarly poor results, with an accuracy of 0.183 and extremely low precision (0.030). Authority involvement may be indicated in a variety of ways—sometimes indirectly—which the model likely fails to grasp. The low recall (0.102) further confirms that the model is not identifying enough relevant instances, perhaps due to unclear definitions of authority involvement in the data.

\begin{figure}[t]
    \centering
    \includegraphics[width=\linewidth]{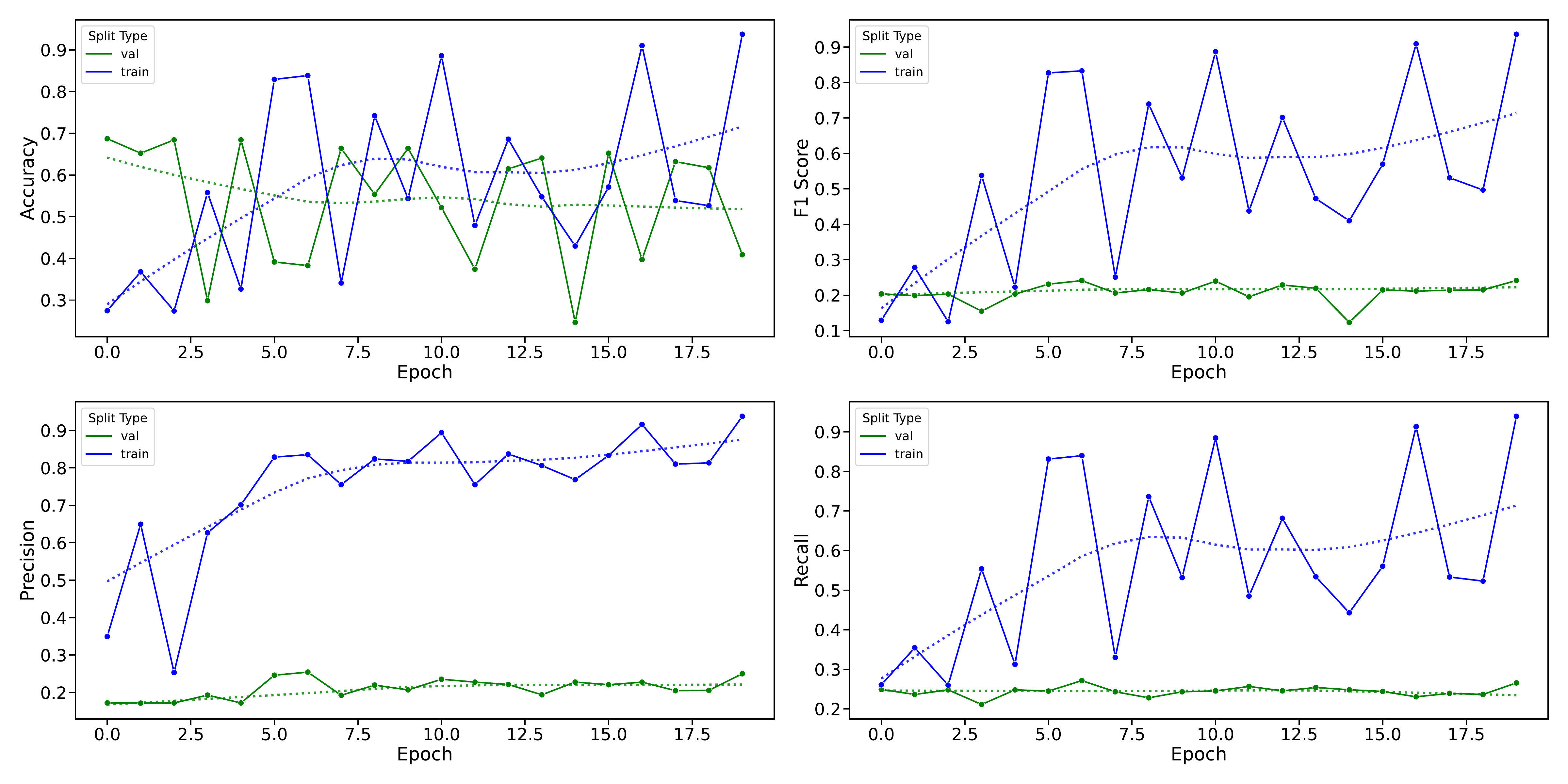}
    \caption{Evaluation metrics while training of the Stance Detection Model}
    \label{fig:training-logs}
\end{figure}

Overall, the model demonstrates varying performance across tasks, with significant discrepancies between accuracy and other key metrics like precision, recall, and F1 score. Accuracy tends to be higher in most cases, but this can be misleading, as it often reflects the model’s tendency to favor majority classes, especially in tasks like News Authenticity Identification, where precision and recall are extremely low despite a high accuracy. This suggests that accuracy alone isn’t a good indicator of success when class imbalance is present.
Tasks such as Political Influence Detection and Scientific Data Usage Detection show relatively balanced performance, with decent F1 scores and alignment between precision and recall. These tasks likely involve clearer patterns or structured data, making them easier for the model to manage. In contrast, tasks like Climate/Environmental Topics Detection and Authority Involvement Detection perform poorly across all metrics, indicating either insufficient or vague data representation for these complex tasks.
The large discrepancy between accuracy and precision/recall in many tasks highlights the model’s difficulty in identifying minority classes. This suggests that improving the model’s sensitivity to underrepresented classes, through better data balancing or more refined feature engineering, is key to boosting its overall effectiveness.

\subsection{Model Training}
Figure \ref{fig:training-logs} shows epoch-wise accuracy, macro precision, recall, and F1 score of training and validation sets. It indicates that while training accuracy and F1 score improve over time, they exhibit high variability, and the validation  metrics remain consistently low, indicating a large performance gap. This suggests the model is overfitting, as it learns the training data well but struggles to generalize to unseen data. Precision on the training set improves steadily, but recall fluctuates, hinting at a possible bias towards the positive class. Despite the rising training metrics, the model's real-world performance, as reflected in the validation scores, shows minimal improvement. It also suggests that training for 10 epochs is enough for modeling.

\begin{table}[t]
    \centering
        \caption{Ablation Studies for the Stance Detection Model.}
    \label{tab:Ablation-Results}
    \begin{tabular}{lccccc}
    \toprule
Hyperparameter & Value & Accuracy ($\uparrow$) & Precision ($\uparrow$) & Recall ($\uparrow$)  & F1 Score ($\uparrow$) \\
\midrule
\multirow{4}{*}{Dropout Rate} & 0.5 & 0.683 & 0.270 & 0.252 & 0.212 \\
 & 0.6 & 0.673 & 0.218 & 0.250 & 0.208 \\
 & 0.7 & 0.664 & 0.223 & 0.251 & 0.215 \\
 & 0.8 & 0.666 & 0.220 & 0.251 & 0.213 \\
\midrule
\multirow{5}{*}{Model Architecture} & 128 & 0.657 & 0.289 & 0.255 & 0.218 \\
 & 256 & 0.651 & 0.232 & 0.252 & 0.216 \\
 & 128,128 & 0.681 & 0.204 & 0.250 & 0.219 \\
 & 128,256 & 0.682 & 0.230 & 0.250 & 0.209 \\
 & 256,256 & 0.687 & 0.210 & 0.250 & 0.207 \\
\midrule
\multirow{2}{*}{Batch Size} & 64 & 0.660 & 0.213 & 0.252 & 0.214 \\
 & 128 & 0.683 & 0.253 & 0.251 & 0.219 \\
\midrule
\multirow{4}{*}{Learning Rate} & 0.005 & 0.676 & 0.225 & 0.251 & 0.212 \\
& 0.001 & 0.652 & 0.270 & 0.253 & 0.217 \\
 & 0.01 & 0.681 & 0.213 & 0.252 & 0.211 \\
 & 0.1 & 0.677 & 0.223 & 0.249 & 0.208 \\
\bottomrule
    \end{tabular}
\end{table}

\subsubsection{Hyperparameter Tuning} \label{sec:grid-ablation}
The ablation study presented in Table \ref{tab:Ablation-Results} provides valuable insights into the effects of various hyperparameters on the performance of the stance detection model. Given the challenge of class imbalance in our dataset, the F1 Score is prioritized for evaluation and selection. The F1 Score balances both precision and recall, making it particularly relevant in scenarios where the distribution of classes is skewed. This focus ensures that we do not solely prioritize overall accuracy at the expense of effectively identifying minority classes, which is crucial for the tasks at hand. We used a grid-search method\footnote{In this grid search, we have trained $4 (\text{Dropout Rate}) \times 5 (\text{Model Architecture}) \times 2  (\text{Batch Size}) \times 4  (\text{Learning Rate}) = 160$ different models, each for 10 epochs.} \citep{lavalle2004relationship} to identify the best hyperparameters for the stance detection task. In this approach, each hyperparameter is systematically varied while keeping the others fixed, allowing us to evaluate the model’s performance across different configurations. As a result, the ablation study table for each hyperparameter presents average values of performance metrics across all tests, where only the specific hyperparameter under evaluation remains stable while the others are changed. 

\textbf{Dropout Rate.} \quad
In Table \ref{tab:Ablation-Results}, we observe varying performance across different values for different dropout rates. Dropout rate of 0.5 yields the highest accuracy (0.683), but the precision and recall metrics are relatively low, indicating that while the model is correct overall, it may struggle to identify true positive instances. As the dropout rate increases to 0.6, 0.7, and 0.8, we see a slight decrease in accuracy, with the precision fluctuating around the same value. The F1 Score remains consistently low across these rates, suggesting that higher dropout rates might lead to under-fitting, where the model fails to capture important patterns due to excessive regularization. We have finalized 0.7 as our ideal dropout rate.

\textbf{Model Architecture.} \quad
Results presented in Table \ref{tab:Ablation-Results} indicate that different configurations yield different outcomes. For instance, a simple architecture with 128 units performs slightly worse than configurations with a combination of layers, such as $[128,128]$ and $[128,256]$. Notably, the best-performing architecture is $[256,256]$, achieving an accuracy of 0.687, though its precision and F1 Score are still modest. This suggests that while deeper architectures may help improve overall performance, they might not sufficiently address the challenges posed by class imbalance, as indicated by the low precision and recall values across all configurations. Following this, we have finalized $[128, 128]$ as our final model architecture, taking the middle, to avoid overfitting.

\textbf{Batch Size.} \quad
Results presented in Table \ref{tab:Ablation-Results} shows that a batch size of 128 shows a better F1 Score (0.219) compared to a batch size of 64. This indicates that larger batch sizes may provide more stable gradient estimates, allowing the model to learn more effectively during training. However, the improvement is modest, implying that while batch size is a relevant factor, it is not the primary driver of performance in this context. Following the results, we have finalized $128$ as our ideal batch size.

\textbf{Learning Rate.} \quad
Results presented in Table \ref{tab:Ablation-Results} indicate that a learning rate of 0.005 yields a notable accuracy of 0.676, with precision and recall metrics remaining stable. Higher learning rates, such as 0.1, show a decrease in F1 Score, which suggests that excessively large steps in parameter space could lead to overshooting the optimal solution. Conversely, the lower learning rate of 0.001, despite being effective in stabilizing the training process, results in lower overall accuracy. We have finalized $0.001$ as our learning rate, focusing on F1 score.

\subsubsection{Impact of Distributional Alignment}
\begin{figure}[t]
    \centering
    \includegraphics[width=\linewidth]{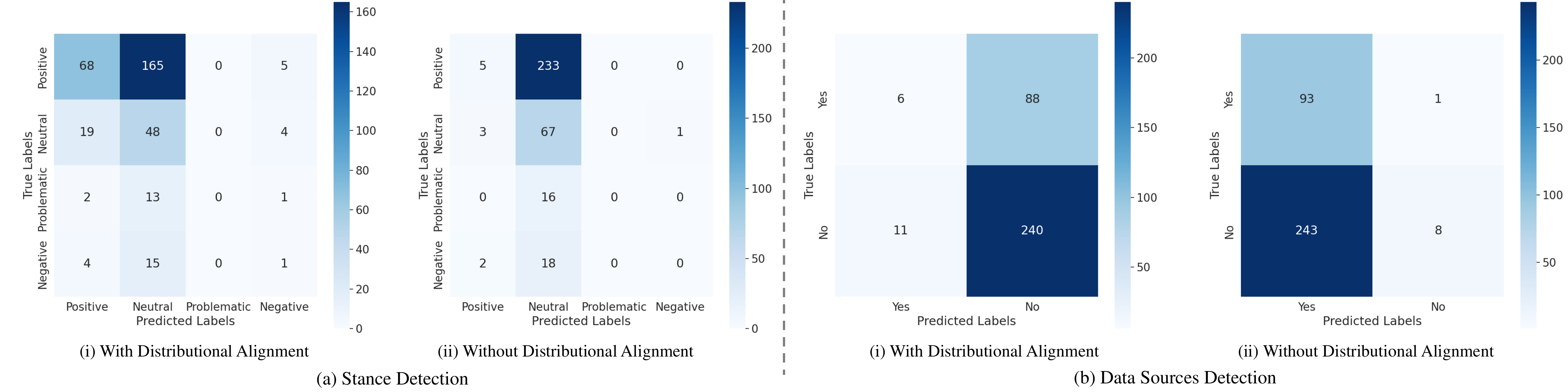}
    \caption{Confusion plots with and without Distributional Alignment.}
    \label{fig:DAConfusionPlots}
\end{figure}

The confusion matrices presented in Figure \ref{fig:DAConfusionPlots} compare model performance with and without distributional alignment for stance detection and data sources detection tasks. For stance detection, the model with alignment (left matrix) shows better accuracy, especially in distinguishing between positive and neutral stances, with fewer misclassifications. Without alignment (right matrix), misclassifications are more frequent, particularly between similar stances, highlighting the benefits of alignment. Distributional alignment reduces bias, enhances generalization, and improves overall accuracy by aligning the training data with real-world distributions. It also helps handle class imbalance and reduces overfitting, ensuring more robust performance across domains. For data sources detection, while alignment slightly reduces misclassifications, the improvement is less pronounced compared to stance detection, but it helped there to move the overall prediction distribution center from 'Yes' to 'No', which seems to be more aligned with the data.

\section{Discussion} \label{sec: discussions}
As awareness of environmental issues increases among governments and the public \citep{Liu2023}, our work with the \texttt{Dhoroni} dataset and the \texttt{BanglaBERT-Dhoroni} model family contributes significantly to this vital discourse. By providing a comprehensive resource that captures climate-related narratives in Bengali, we aim to facilitate informed discussions on sustainable environmental management. Our efforts not only empower environmental managers and stakeholers, but also engage a wider audience, enabling them to better understand and respond to the complexities of climate change \citep{Ghimire2016} and its local implications.

\subsection{Societal Impact}
Societal impact of the \texttt{Dhoroni} dataset cannot be overstated. In Bangladesh, where climate change poses significant threats \citep{Nishat2013}, understanding public discourse around these issues is crucial for driving social change \citep{Bell1994}. The dataset offers a platform for analyzing how climate-related information is communicated and perceived within the community. This understanding can empower citizens to advocate for more robust climate policies and hold leaders accountable for environmental actions.

By providing a rich repository of news articles, the dataset encourages grassroots movements focused on climate awareness and action. Local organizations can harness insights derived from the dataset to mobilize communities, fostering a collective sense of responsibility toward environmental conservation. This grassroots approach can create a ripple effect, leading to increased public engagement and advocacy for climate-related policies.
Furthermore, the dataset's emphasis on authenticity and data sources plays a pivotal role in combating misinformation. In an age where misinformation can easily spread through social media and news outlets \citep{Allcott2019}, having a reliable dataset to verify claims becomes essential. By highlighting credible sources and assessing the authenticity of information, the dataset can serve as a foundation for educating the public about the importance of fact-checking and responsible consumption of news \citep{LpezMarcos2021}.

\texttt{BanglaBERT-Dhoroni} family models can significantly help in addressing societal issues by improving the detection of key elements in news, such as stance, political influence, and authority involvement. By accurately identifying these factors, the models can help assess the objectivity of news reports and provide insights into how political or institutional biases may shape public discourse \citep{CHOULIARAKI2000}, fostering a more informed society.
In combating false news, the models for detecting scientific and statistical data usage, as well as data sources, play a crucial role \citep{Mishra2022}. They can help verify the credibility of the information, ensuring that data presented in news is properly sourced and authentic. This can reduce the spread of misinformation, particularly on important topics like climate change, health, and political affairs.

Societal implications of the dataset also extend to gender and social equity issues. Climate change disproportionately affects marginalized communities, including women \citep{qwfqf2017}, low-income households \citep{Haque2014}, and indigenous populations \citep{sgaeagas2013}. By analyzing media representations of these groups in the context of climate change, researchers can identify gaps in coverage and advocate for more inclusive narratives. This focus on equity is essential for ensuring that all voices are heard in the conversation about climate change and its impacts.
Ultimately, the \texttt{Dhoroni} dataset has the potential to catalyze positive societal change by fostering informed discussions, encouraging activism, and promoting equity in climate discourse. By leveraging the insights generated from this dataset, individuals and organizations can work together to address climate challenges and create a more sustainable future.

\subsection{Global Impact}
Global impact of the \texttt{Dhoroni} dataset extends beyond the borders of Bangladesh, contributing to the broader conversation on climate change and environmental issues. As climate change is a global challenge \citep{GAEWGAHG2023}, understanding how different countries and cultures discuss these topics is vital. By providing a dataset focused on Bengali perspectives, the \texttt{Dhoroni} project adds an important dimension to the global discourse, highlighting the voices and experiences of a population that is often overlooked in international discussions.

The inclusion of Bengali-language content in global climate research is crucial for creating a more balanced narrative. Much of the existing climate-related research and discourse is dominated by English-language sources \citep{Ghosh2019}, which can skew the understanding of how climate change affects various regions and communities \citep{LealFilho2022}. By amplifying Bengali voices, the dataset helps to ensure that the unique challenges and perspectives of Bangladeshis are recognized in global climate discussions.
Additionally, the dataset can facilitate cross-cultural comparisons in climate reporting and public perception. Researchers and policymakers can analyze how climate change is framed in Bangladesh compared to other countries, revealing potential differences in priorities and narratives. This comparative analysis can inform international collaboration on climate initiatives, allowing for a more holistic understanding of the global climate landscape.
On a global scale, \texttt{BanglaBERT-Dhoroni} family models can enhance understanding of how different regions are affected by news topics, such as environmental impacts or political issues. By detecting authenticity and target groups, they can help gauge the global relevance of news stories \citep{Gurevitch1994} and ensure that critical issues, such as authority involvement and environmental topics, are accurately represented across different parts of the world.

Moreover, as global climate initiatives continue to evolve, the \texttt{Dhoroni} dataset can serve as a benchmark for measuring progress in climate communication. By tracking changes in media representation and public sentiment over time, researchers can assess the effectiveness of climate campaigns and policies. This ability to evaluate the impact of interventions is crucial for refining strategies and ensuring that efforts to combat climate change are informed by evidence-based insights.
In essence, the \texttt{Dhoroni} dataset plays a pivotal role in connecting local narratives to global conversations about climate change. By showcasing the importance of Bengali perspectives, it contributes to a more inclusive and equitable approach to addressing environmental challenges on a global scale.

\subsection{\textit{Dhoroni} in Environmental Policy-making and Management}
The \texttt{Dhoroni} dataset is a significant advancement in NLP for environmental management, especially for Bengali-speaking communities. By curating 2,300 climate-related news articles, it offers insights into how local media frames environmental issues, which is crucial for informed policymaking and public engagement  \citep{Sofiullahi2024}. The fine-tuned \texttt{BanglaBERT-Dhoroni} models, leveraging transformer-based architectures, facilitate accurate detection of climate topics, authority involvement, and media perspectives. These insights help stakeholders address misinformation, promote awareness, and ensure marginalized voices are represented \citep{Chen2019}. Overall, the \texttt{Dhoroni} dataset and model family provide a powerful resource to enhance understanding of climate issues, empowering communities to proactively address climate challenges.

\subsection{Limitations}
While the \texttt{Dhoroni} dataset represents a significant advancement in the study of climate change discourse in Bengali, it is essential to acknowledge its limitations. The dataset is limited to news articles, which may not capture the full spectrum of climate-related discussions occurring in other formats, such as social media, blogs, and community forums. While news articles are valuable for understanding mainstream narratives, they may miss the grassroots perspectives that are crucial for a holistic understanding of climate change impacts. Future research could benefit from incorporating a broader range of data sources to capture diverse voices and narratives.
Lastly, the dynamic nature of climate change discourse means that the dataset may quickly become outdated. New developments in climate science, policy, and public sentiment can significantly alter how climate issues are discussed in the media. Regular updates to the dataset will be necessary to ensure its continued relevance and utility for researchers and practitioners.
Despite these limitations, the \texttt{Dhoroni} dataset represents a substantial step forward in understanding climate change discourse in Bengali. By addressing these limitations in future research, we can enhance the dataset's impact and continue to support the critical study of climate change communication.

\subsection{Future Work Directions}
For future work, we propose two broad areas: utilizing Large Language Models and improving Low-Performing Models.

\subsubsection{Training LLMs on Dhoroni}
The advent of large language models (LLMs) has transformed the field of natural language processing \citep{https://doi.org/10.48550/arxiv.2304.02020}, offering powerful tools for understanding and generating human language \citep{Zhou2023}. Training LLMs on the \texttt{Dhoroni} dataset presents a significant opportunity to advance the capabilities of NLP for the Bengali language, particularly in the context of climate change and environmental issues. By leveraging the rich data contained within the \texttt{Dhoroni} dataset, researchers can develop LLMs that are fine-tuned to recognize and generate Bengali text related to climate topics.

One of the primary benefits of training LLMs on this dataset is the potential for enhanced performance in various NLP tasks. For example, stance detection models fine-tuned on the \texttt{Dhoroni} dataset could become highly proficient at understanding public sentiment regarding climate change, enabling more effective communication strategies for climate-related campaigns. Additionally, LLMs can be applied to extract insights from the dataset \citep{https://doi.org/10.48550/arxiv.2304.00477}, summarizing key themes and trends in climate discourse over time.
Moreover, training LLMs on the \texttt{Dhoroni} dataset can contribute to the development of applications that directly address climate challenges. For instance, chatbots powered by these models could engage with the public on climate issues, providing information, resources, and guidance on sustainable practices. These tools can empower individuals to make informed decisions and take action in their communities, fostering a sense of collective responsibility toward environmental stewardship.

Furthermore, LLMs trained on the \texttt{Dhoroni} dataset can facilitate better information retrieval and recommendation systems for climate-related content. By understanding the nuances of climate discourse in Bengali, these models can enhance the accessibility of relevant information for researchers, policymakers, and the general public. Improved information retrieval can help bridge the gap between scientific research and public understanding, ultimately leading to more informed decision-making in addressing climate change.
In addition to practical applications, training LLMs on the \texttt{Dhoroni} dataset can also contribute to advancing the state of NLP research for low-resource languages. Bengali is often considered a low-resource language in terms of NLP tools and resources \citep{Banik2019}, and developing robust LLMs for this language can serve as a model for similar efforts in other low-resource languages. By demonstrating the effectiveness of fine-tuning LLMs on datasets like \texttt{Dhoroni}, researchers can encourage the development of similar initiatives for other underrepresented languages.

\subsubsection{Performance Improvement in Low-Performing Models}
To improve poor performance in tasks such as detecting climate / environmental topics, detecting news targets, and detecting involvement of the authority, several modeling approaches can be explored. These tasks involve complex patterns that current methods may struggle to capture.
Multi-task learning can enhance performance by training related tasks together, allowing models to share learned features. For example, training in news target detection and authority involvement detection simultaneously may help the model generalize better. Lastly, further hyperparameter tuning and regularization techniques, such as learning rate schedulers, dropout, and weight decay, can optimize model performance and reduce overfitting. Ensembling multiple models may also improve robustness by leveraging the strengths of different approaches for better overall performance.

\section{Conclusion}
In conclusion, introducing \texttt{Dhoroni} dataset marks a significant advancement in the field of natural language processing for Bengali, particularly in the context of climate change and environmental discourse. With its comprehensive collection of 2,300 Bengali news articles, this dataset provides a rich resource for researchers and practitioners alike. The rigorous annotation process, conducted by three annotators across ten different perspectives, ensures a multi-faceted approach to analyzing climate-related issues. Each perspective is accompanied by detailed reasoning, enhancing the dataset's usability and allowing for deeper insights into the complexities of climate narratives in the Bengali media.
\texttt{BanglaBERT-Dhoroni} model family provides a strong framework for tackling various tasks in the *Dhoroni* dataset, with stable benchmarking scores demonstrating the effectiveness of fine-tuning. This foundation supports future research and highlights the importance of developing NLP resources for low-resource languages like Bengali. Overall, the \texttt{Dhoroni} dataset and its models mark a significant step toward deepening our understanding of climate discourse in Bengali, fostering inclusive conversations, and contributing to global efforts to address environmental challenges. We anticipate valuable insights from further exploration of this dataset to drive action and raise awareness within the Bengali-speaking community and beyond.

\section*{Acknowledgement}
The authors thank the Computational Intelligence and Operations Laboratory (CIOL) for their support, as well as for the initial review and feedback on the manuscript.

\section*{Author Contributions}
A.T.W. and W.F. contributed to the work equally.
\textbf{A.T.W.:} Conceptualization of this study, Methodology, Data curation, Resources, Investigation, Formal analysis, Supervision, Validation, Visualization, Writing – original draft, Writing – review \& editing, Project administration. 
\textbf{W.F.:} Software, Methodology, Formal analysis, Investigation, Resources, Data Curation,
Project administration. 
\textbf{T.A.:} Resources, Data Curation. 
\textbf{A.R.:} Resources, Data Curation, Writing – original draft, Writing – review \& editing. 
\textbf{M.R.I.:} Resources, Data Curation, Writing – review \& editing.

\section*{Competing Interests}
The authors declare that there are no competing interest regarding the publication of this paper. All research procedures followed ethical guidelines, and the study was conducted with integrity and transparency. The authors have no financial, personal, or other relationships that could inappropriately influence or bias the content of this work.

\section*{Ethics Statement}
All news articles in this dataset have been collected in accordance with ethical guidelines and journalistic standards. The anonymity of the newspapers and media outlets has been preserved to protect their identity. Some scientific information in the articles includes citations from government or relevant authoritative bodies, which are not anonymized to ensure the authenticity of the dataset. Harmful or defamatory content has been removed to maintain a respectful and safe dataset. Accuracy and diversity have also been prioritized in curating the news to reflect a balanced and reliable collection of information.

\section*{Data Availability}
Data related to this work is publicly available in \texttt{Zenodo} at \textit{\href{https://doi.org/10.5281/zenodo.13695110}{\texttt{DOI: 10.5281/zenodo.13695110}}} \citep{wasi_2024_13695110} under the MIT license.

\section*{Materials Availability}
Trained models are publicly available as \texttt{Collection} in \texttt{Hugging Face} at \textit{\href{https://huggingface.co/collections/ciol-research/dhoroni-6723c1c675c6ef21285c0dda}{\texttt{ID: ciol-research/dhoroni-6723c1c675c6ef21285c0dda}}} under the MIT license.

\section*{Funding Statement}
This work received no specific grant from any funding agency, commercial or not-for-profit sectors.

\bibliography{iclr2025_conference}
\bibliographystyle{iclr2025_conference}


\end{document}